%% file: SEMpaper.tex
\documentclass[final]{article}
\usepackage[utf8]{inputenc}
\usepackage[margin=1.3in]{geometry}
\usepackage{mathptmx}
\usepackage{graphicx}
\usepackage{color}
\usepackage{amsfonts}       
\usepackage{nicefrac}       
\usepackage{microtype}      
\usepackage{amsmath,graphicx}

\usepackage{amsmath}
\usepackage{siunitx}

\usepackage{outlines}
\usepackage[ruled]{algorithm2e}

\graphicspath{{figs/}{./}}

\newcommand{\BK}[1]{\textcolor{black}{#1}}
\newcommand{\SL}[1]{\textcolor{black}{#1}}

\title{Explainable Deep Learning for Uncovering Actionable Scientific Insights for Materials Discovery and Design}

\author{
Shusen Liu, Bhavya Kailkhura, Jize Zhang\\
Center for Applied Scientific Computing, Computing Directorate \\
Lawrence Livermore National Laboratory \\
\texttt{liu42, kailkhura1, zhang64@llnl.gov} \\
\and
Anna M. Hiszpanski, Emily Robertson, Donald Loveland, T. Yong-Jin Han \\
Materials Science Division, Physical and Life Science Directorate \\
Lawrence Livermore National Laboratory \\
\texttt{hiszpanski2, robertson37, loveland4, han5@llnl.gov} \\
}

\begin{document}
\maketitle

\begin{abstract}

The scientific community has been increasingly interested in harnessing the power of deep learning to solve various domain challenges. However, despite the effectiveness in building predictive models, fundamental challenges exist in extracting actionable knowledge from deep neural networks due to their opaque nature. In this work, we propose techniques for exploring the behavior of deep learning models by injecting domain-specific actionable attributes as tunable ``knobs'' in the analysis pipeline. By incorporating the domain knowledge in a generative modeling framework, we are not only able to better understand the behavior of these black-box models, but also provide scientists with actionable insights that can potentially lead to fundamental discoveries.

\end{abstract}

\input{intro.tex}

\input{related.tex}

\input{background.tex}

\input{method.tex}
\input{results.tex}

\input{discussion.tex}

%

\section*{Acknowledgement}
This work was performed under the auspices of the U.S. Department of Energy by Lawrence Livermore National Laboratory under Contract DE-AC52-07NA27344. This work is reviewed and released under LLNL-JRNL-811201.
We would also like to extend our gratitude and appreciation to Brian Gallagher for his valuable feedback and comments on the manuscript.

\bibliographystyle{unsrt}
\bibliography{refs}

\newpage
\input{supplemental.tex}

\end{document}

%% file: intro.tex
\section{Introduction}

Due to the tremendous success of deep learning in commercial applications, there are significant efforts to leverage these tools to solve various scientific challenges. Deep learning automatically discovers a suitable feature representation from the raw data that allows powerful predictive models to be built from large and complex datasets. Unfortunately, this benefit comes with a major limitation – these complex models are often considered as black-boxes, and understanding or explaining their inner workings is extremely difficult.

Besides the inherent complexity of deep learning models, their application to the scientific domain also has unique challenges compared to the traditional applications in commercial domains. Scientific data often requires domain knowledge to be understood and annotated, which often leads to label sparsity. Furthermore, instead of focusing on the predictive performance, in scientific applications, we particularly value the insights distilled from the model that can potentially advance our scientific understanding.

\begin{figure*}[t]
\centering
\vspace{-4mm}
  \includegraphics[width=0.98\linewidth]{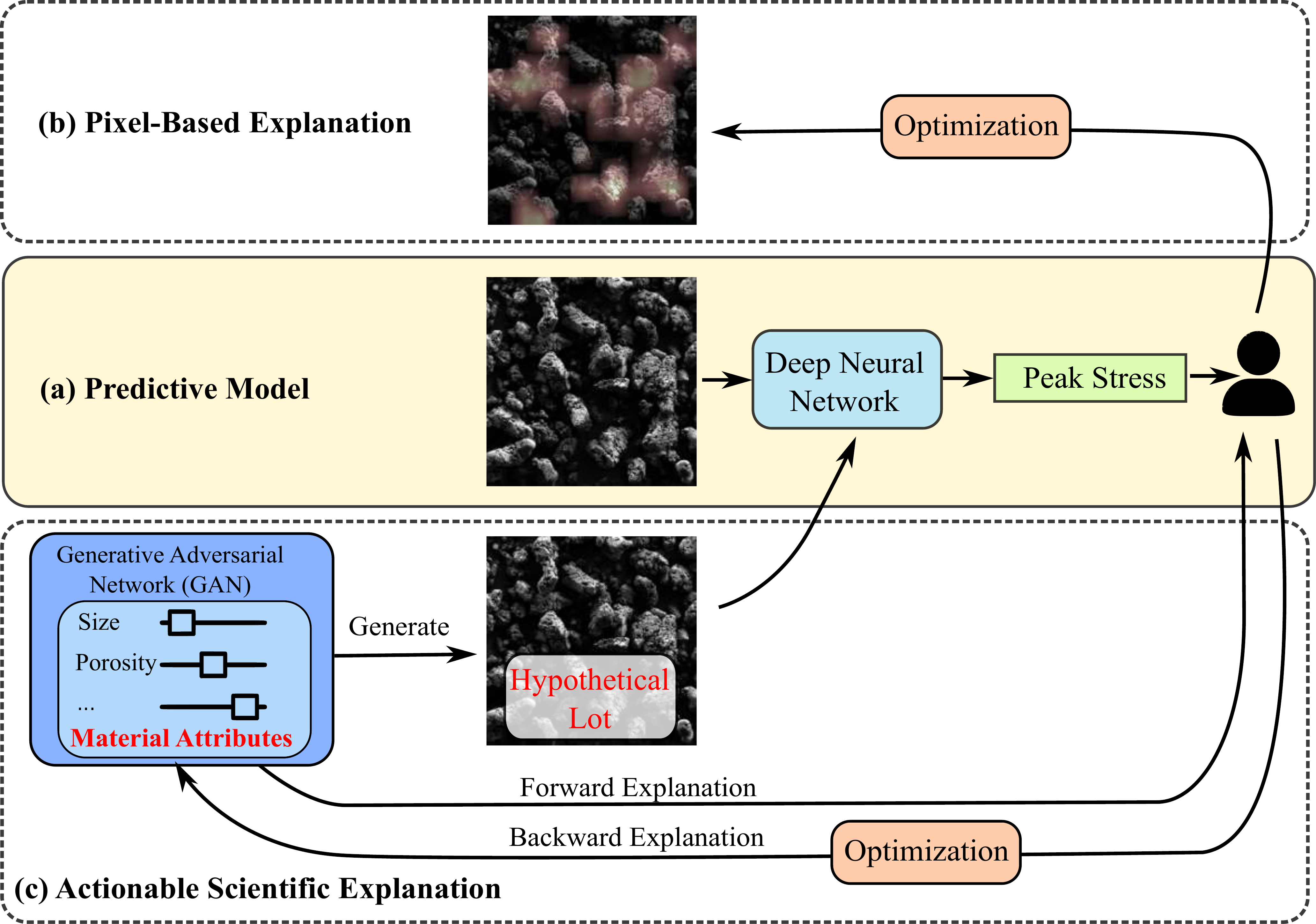}
 \caption{
Overview of the actionable explanation pipeline. We have a deep neural network model (a) for predicting material peak stress from SEM images. Instead of trying to attribute the decision to the input pixel space (e.g., GradCAM~\cite{selvaraju2017grad}) (b), which cannot produce understandable and actionable solution, we can relying on a generative model to produce a hypothetical lot that is conditioned on the key attributes of the material, from which we can obtain an explanation that is not only directly understandable by the material scientist but also can easily be translated into actionable guidelines in the material synthesis process (c).
}
\label{fig:pipeline}
\end{figure*}

Many existing scientific applications of deep learning focus on building a predictive model for certain experimental output modality (e.g., building a model for predicting the material peak stress given a scanning electron microscope image~\cite{gallagher2020predicting}).
However, despite their effectiveness in predicting the quantity of interest, we do not have a viable way to evaluate and reason about their decisions to the domain scientists.
However, despite their effectiveness in predicting the quantity of interest, we do not have a viable way to evaluate and reason about their decisions to the domain scientists. Even if we believe the model accurately captures the underlying scientific principle, the model opacity makes it extremely hard to extract useful information from the model that can be turned into actionable insights for discovery. One key reason that leads to these challenges is our inability to reason about domain attributes in the deep learning pipeline that are meaningful to the scientists.
To motivate the role of domain attributes in understanding the model behavior, we present a real-world application of model understanding in material science.

\noindent\textit{\textbf{Motivating Example:}}
As illustrated in Figure~\ref{fig:pipeline}(a), let us assume that we have a deep learning model that predicts the peak stress of the material given a scanning electron microscope (SEM) image as an input.
The traditional pixel-based explanation approaches~\cite{ZeilerFergus2014, bach2015pixel,selvaraju2017grad} for the convolutional neural network (CNN) produces a heat-map (on a per-pixel level) to highlight the region in the image that contributed the most to the prediction. Such an approach may work well for natural images, e.g., highlighting the head of the cat when predicting a cat image.
However, this per-pixel explanation is not particularly insightful when trying to explain why a certain material has a higher peak stress by highlighting pixels in the image as illustrated in Figure~\ref{fig:pipeline}(b). The reason for this lack of insight being that the image pixel space does not correspond to any meaningful or understandable material science concepts. Furthermore, a material scientist may be more interested in understanding the effect of only a subset of all possible attributes that are explicit and actionable (e.g., crystal size, etc.).

In this work, we aim to address this fundamental explainability challenge by injecting domain attributes in a post-hoc manner into the prediction pipeline by utilizing advances in deep generative modeling~\cite{goodfellow2014generative}.
As illustrated in Figure~\ref{fig:pipeline}, we first build a generative model that can produce ``fake (or hypothetical)'' SEM images compliant to user-controlled attributes, e.g., an SEM image of a hypothetical material with a larger or smaller average crystal size than a given reference material.
We then leverage these attributes as the explainable handles to reason more effectively by probing the predictive model behavior with generated hypothetical materials. This approach allows us to answer the questions in the language that the domain scientists understand, i.e., \textit{how does changes in the crystal size (or porosity, etc.) impact the peak stress prediction? or how should material attributes be altered to obtain a material with higher peak stress?}
\SL{
Note that compared to the correlation analysis between material attributes and prediction outputs, the proposed method not only produce a per-instance explanation but also generates the corresponding hypothetical SEM image that reflects the manifestation of optimal attribute changes to reach a certain objective, e.g., higher peak stress.
Such images of hypothetical materials can be particularly helpful to the material scientists for gaining intuitive understanding for what type of material should be targeted during synthesis to attain the desired properties and potentially revealing other previously unknown variations that is not captured by already known attributes.}

A crucial component for the success of such an explanation scheme is the ability to generate appropriate images corresponding to given changes in the attribute values. However, training a generative model to generate high-quality images conditioned on given attribute values has proven to be challenging. In this work, we solve this problem by adopting an image editing model rather than generating images from scratch. Specifically, we take an image and target attributes as inputs, and then perform selective editing of the desired attributes in the given image. The additional meta-information of an input image allows us to train high-quality editing models capable of generating hypothetical images that capture intricate details of the material attributes and are indistinguishable from real SEM images.
Incorporating domain attributes in the generative modeling pipeline allows us to better understand the behavior of black-box predictive models through the perspective of generated hypothetical materials.
Further, it provides scientists with actionable insights that can potentially lead to new discoveries.

The key contributions of our work are listed as follows:
\begin{itemize}
	\item We propose an explainable deep learning approach to provide {\em actionable} scientific insights;
	\item We demonstrate that the generative model crucial for our approach can capture the association between domain attributes and intricate image features with extremely small amount of supervised information;
	\item We showcase the usefulness of the proposed approach in a real-world application of feedstock material synthesis by providing domain scientists with actionable insights to improve the  material quality.
\end{itemize}

%
%

%% file: related.tex
\section{Related Works}

With the recent advances in deep learning, scientists are increasingly relying on data-driven modeling for solving scientific challenges.
Machine learning has been successfully applied to a variety of domains, such as physics~\cite{peterson2017zonal, anirudh2019improved}, biology~\cite{webb2018deep}, material science~\cite{butler2018machine}, and many more
\cite{reichstein2019deep, kurth2018exascale, baldi2001bioinformatics}.
Many of these scientific machine learning applications either focus on building accurate surrogate models for the underlying physical phenomenon or developing sophisticated predictive models from complex simulation/experimental output (e.g., image and time-series) to predict various properties of interest.
Furthermore, few existing works have looked into addressing the explainability challenges in the context of scientific applications~\cite{SimonyanVedaldiZisserman2013, ZeilerFergus2014, YosinskiCluneNguyen2015, bach2015pixel, lapuschkin2019unmasking, kailkhura2019reliable}.

In machine learning research, the opaque nature of deep neural networks has prompted many efforts to improve their explanation and interpretation~\cite{xie2020explainable}.
Most of these works focus on traditional computer vision or natural language processing applications.
One key strategy for interpretation is attributing the prediction importance into the model's input domain, most notably for the convolution neural network (CNN).
Various approaches~\cite{SimonyanVedaldiZisserman2013, ZeilerFergus2014, YosinskiCluneNguyen2015, bach2015pixel, lapuschkin2019unmasking} have been proposed to highlight the important regions of an input image that contributes the most to a decision.
One can also consider the attribution scheme from a model agnostics perspective~\cite{RibeiroSinghGuestrin2016, KrausePererNg2016, LundbergLee2017}, e.g., the LIME~\cite{RibeiroSinghGuestrin2016} explains a prediction by fitting a localized linear model for approximating the classification boundary for a given prediction.
However, such an attribution is only useful if the input domain itself is explainable to the user (e.g., natural images), which is not the case in many scientific applications (e.g., complex SEM images).
Moreover, the assignment of the importance to input (or pixel) space is limited in the sense that it can only provide passive correlative information.
Many important insights can only be obtained from a counterfactual understanding involving interventional operation, i.e., what kind of \emph{changes} to the input is necessary if a specific model output is requested.
To address these challenges counterfactual explanation   approaches~\cite{kusner2017counterfactual,narendra2018explaining, Goyal2019, anne2018grounding} have been proposed, e.g., the counterfactual visual explanation~\cite{Goyal2019} work introduces a patch-based image editing and optimization scheme for obtaining interpretable changes in the pixel space for altering a prediction.
However, the patch-based editing can severely limit the expressiveness of the input modification.
In this work, we use a generative adversarial neural network (GAN) for making meaningful input interventions leveraging GAN's latent space. Specifically, by leveraging the GAN model that is able to meaningfully edit the model inputs (e.g., image) via domain-specific attributes, we allow explanation in the context of the application for obtaining actionable domain understanding.
\SL{GAN-based explanation of a classifier was explored and applied to non-scientific applications in our preliminary work~\cite{liu2019generative} in a similar direction. In this paper, our focus is however on regression problems for scientific applications.}

\SL{
On the material science front, the utilization of machine learning for predicting material properties has been investigated in multiple previous works~\cite{decost2017characterizing, webel2018new, gallagher2020predicting}. In particular, predicting peak stress from SEM images has been studied in the work by Gallagher et al.\cite{gallagher2020predicting}, in which both traditional computer vision techniques and deep neural network approaches have been explored.
Compared to the state-of-the-art peak stress prediction works that focus on making accurate predictions, the proposed work focuses on how to extract actionable scientific insights by explaining what features in images are important to the presumably accurate prediction of these models.}
Although we focus on a specific materials application in this paper, the proposed actionable explanation generation approach is quite generic and can be applied to a wide range of scientific applications.

%% file: background.tex
\section{Background}
\label{sec:background}
In this work, we demonstrate how the proposed explanation scheme can help obtain actionable insights in a material science application. Specifically, we are interested in understanding the behavior of a deep learning model that was trained on SEM images of feedstock materials for predicting their respective mechanical properties.

Feedstock materials are basic building blocks for producing increasingly sophisticated components, prototypes, or finished products. These materials are often optimized to meet certain performance requirements before they can be appropriately utilized. One persistent challenge originated from developing and deploying materials in a timely manner is the significant time and resources required to optimize the material to meet the desired specification. In material science applications, we hope to accelerate the material development process by leveraging the modeling capability of deep learning on increasingly complex and heterogeneous experimental data. In particular, by learning the relationship between the salient feature of observed data (e.g., SEM images) and the material's characteristics, the model can provide valuable feedback to deepening the scientists' understanding, which in turn will help accelerate the material design and optimization processes.

\begin{figure*}[htbp]
\centering
  \includegraphics[width=0.99\linewidth]{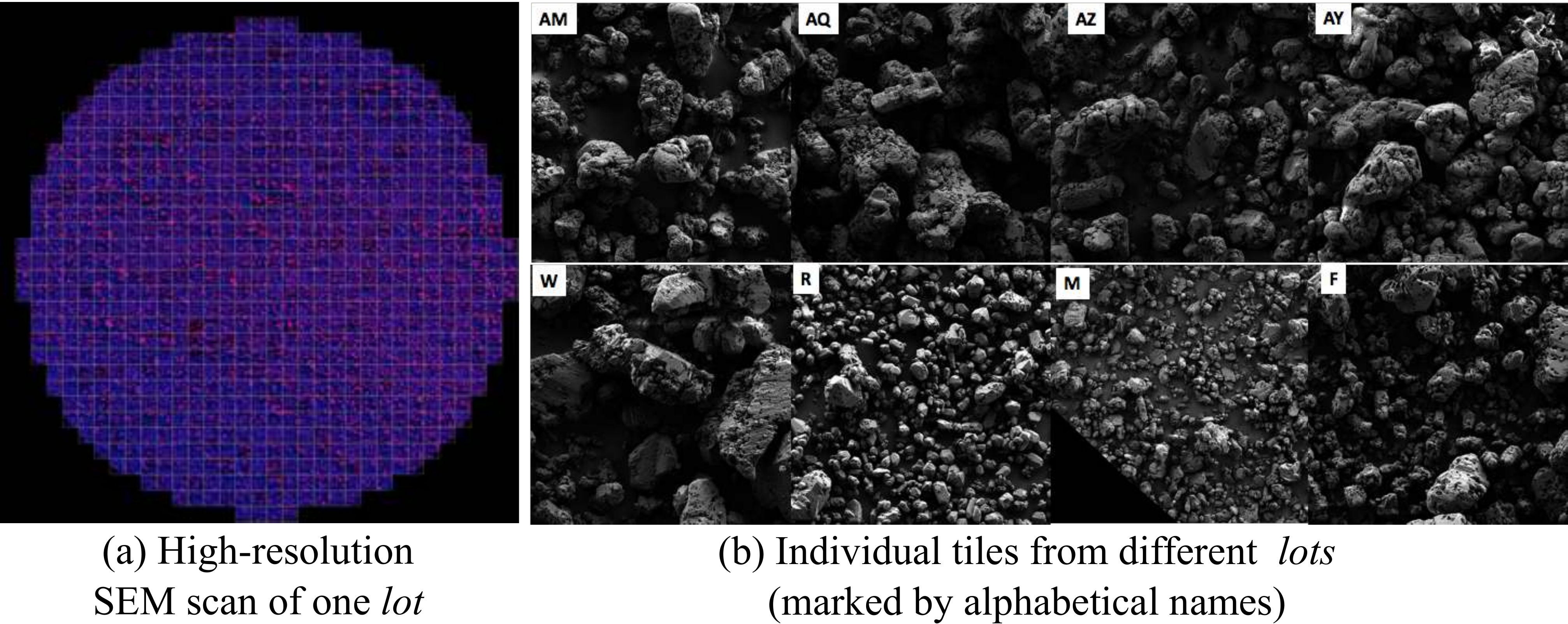}
 \caption{
  The SEM high-resolution scan (a) of a given lot is divided into smaller image tiles. All the image tiles from 30 lots is used for training a CNN-based peak-stress prediction network. In (b), examples of the tile from different lots are illustrated. We can see that each image captured key characteristic, e.g., crystal size, of their respective lots.
}
\label{fig:SEMScan}
\end{figure*}

%
In our exemplar case study, the feedstock material of interest is 2,4,6-triamino-1,3,5- trinitrobenzene (TATB) and its compressive strength upon compaction. The compressive strength of compacted TATB can vary significantly with changes in the TATB’s crystal characteristics, including average size, size distribution, porosity and surface textures to name a few.  The experiment involves 30 different synthesis batches (referred to as lots in the context of this work) of material samples, with each batch showing different overall crystal characteristics. Each of the 30 lots is analyzed with a Zeiss Sigma HD VP scanning electron microscope (SEM) using a \SI{30.00}{\micro\metre} aperture, 2.00 keV beam energy, and ca. 5.1 mm working distance to capture high-resolution images. The software Atlas is used to automate the image collection. As illustrated in Figure~\ref{fig:SEMScan}, for each sample, the entire SEM stub surface is mapped, and corresponding images are collected with slight overlap to create a stitched mosaic of the full area.  The field of few per each mosaic tile is \SI{256.19}{\micro\metre} $\times$ \SI{256.19}{\micro\metre} with a pixel size of \SI{256.19}{\nano\metre} $\times$ \SI{256.19}{\nano\metre} ( $1024 \times 1024$ image size).  In total, we captured 69,894 sample images from 30 lots of TATB.
\SL{These images are then down selected by removing the ones with black margins (i.e., at edge of the scan) and other inconsistencies to ensure the quality of the training and validation sets, which consists of 59,690 images.}
To better characterize the images in each lot, two material scientists provided by visual inspection quantitative estimations of several key material attributes, such as crystal size, porosity, size dispersity, and facetness (the detail of these concepts are discussed in Section~\ref{sec:method}).

%
The stress and strain mechanical properties are tested for each lot by uniaxially pressing duplicate samples from each TATB powder lot in a cylindrical die at ambient temperature to 0.5 in. diameter by 1 in. height, with a nominal density of 1.800 g/cc. Strain controlled compression tests were run in duplicate at \SI{23}{\celsius} at a ramp rate of 0.0001 $s^{-1}$ on an MTS Mini-Bionix servohydraulic test system model 858 with a pair of 0.5-inch gauge length extensometers to collect strain data. From the obtained stress-strain curve, only the peak stress values were considered as the outputs of the machine learning models, resulting in an image dataset, in which the same properties are assigned to all images (tiles) from the same lot.
A deep neural network regressor is then trained to predict the stress/strain value from given images (tiles).
Even though the prediction is based on a small patch of the whole SEM scan, the material scientists hypothesize that the individual image should contain salient information that is indicative of the behavior of the entire lot.
Provided the prediction is accurate for unseen lots, such a predictive model is a valuable tool for material scientists to quickly screening candidate materials to prioritize for laboratory testing.  However, despite the ability to down-select potential candidates for further evaluations, the material scientists still need to produce the sample and carry out SEM imaging procedure, which is an extremely time-consuming process.
Furthermore, even though the prediction model appears to capture the relationship between the material image features and their performance, the material scientists cannot directly obtain or reason about such understanding to guide the next set of experiments to perform to quickly obtain the desired materials, i.e., producing material with specific features (e.g., crystal size, porosity) that can potentially lead to better and desired performances.
In this work, we aim to address the challenge of extracting domain insights from the predictive model and provide actionable guidance to the material scientist for the material manufacturing process.


%% file: method.tex
\section{Method}
\label{sec:method}
As illustrated in Figure \ref{fig:pipeline}, the proposed technique includes three key components: 1) the predictive model, 2) the attribute-guided generative model, and 3) the optimization module that leverages the predictive model and image generation model for obtaining actionable scientific insights.

\SL{In this section, we first provide an overview of the predictive model for predicting the compressive strength from SEM images.}
Next, we discuss how we can utilize the attribute conditioned generative adversarial network for generating meaningful image modifications even when we only have extremely sparse labels.
We then introduce two novel modes of explanation relying the image generation pipeline, one following a forward evaluation process, whereas the other relying on the optimization module using gradient backpropagation, to reason about model behavior through domain attributes.

\begin{figure*}[htbp]
\centering
\vspace{-4mm}
  \includegraphics[width=0.9\linewidth]{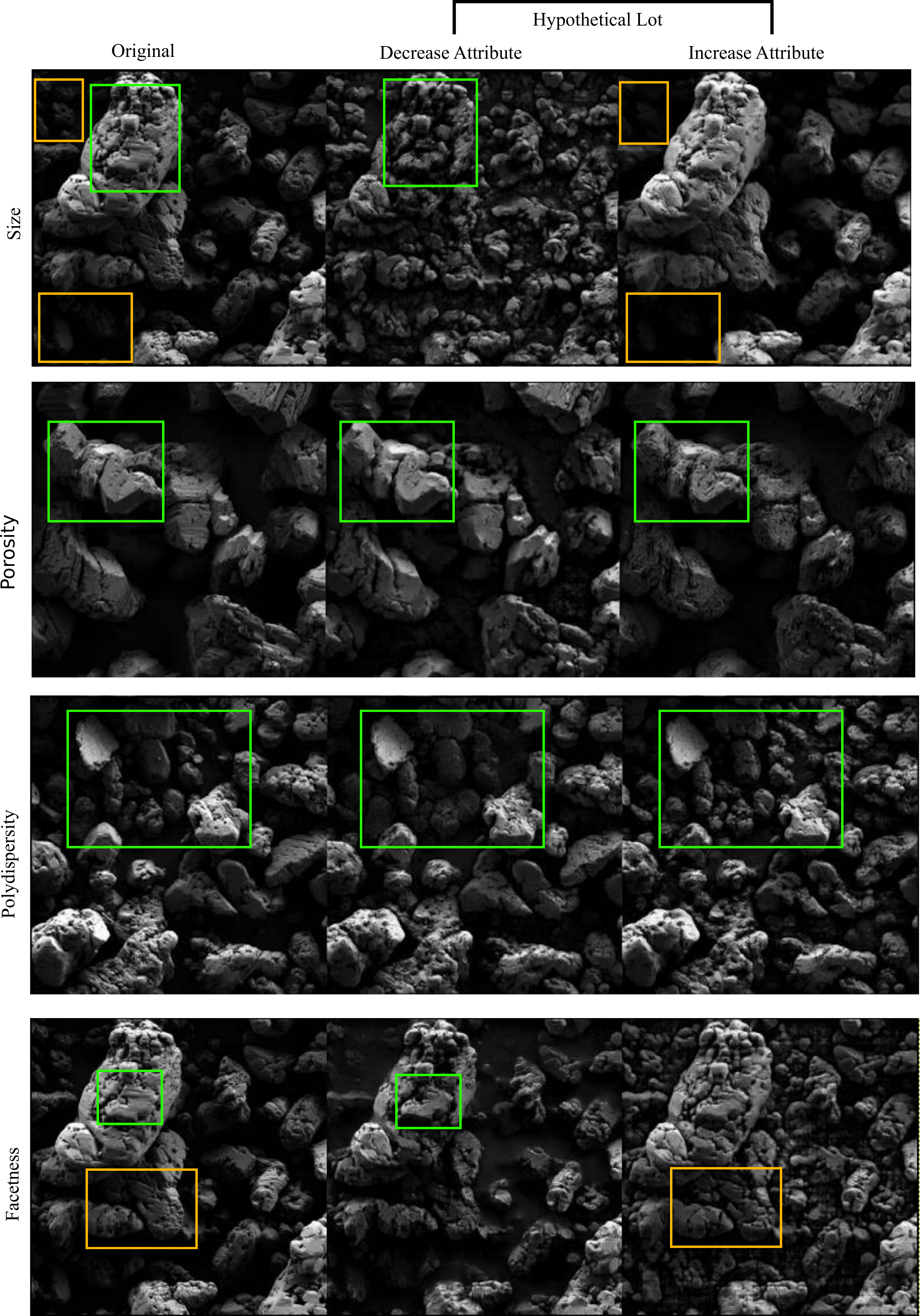}
 \caption{
 Illustration of material attributes-guided SEM image generation. The left column is the original SEM image. The middle and right column show the GAN generated images of hypothetical \emph{Lots} that increase or decrease the corresponding material attributes, respectively. The colored boxes highlight the corresponding regions in the image (different colors mark different regions in the image), in which we can find clear changes that reflect the alterations in the attribution.
}
\label{fig:attEdit}
\end{figure*}

\subsection{Predictive Model}
\label{sec:predModel}
\SL{
As discussed in Section~\ref{sec:background}, a deep neural network regression model is trained to predict the peak stress of a material \emph{Lot} from a given SEM image tile (see details in Figure~\ref{fig:SEMScan}(b)).
The regression model is built upon the WideRestNet CNN architecture~\cite{zagoruyko2016wide} and trained on all 30 \emph{Lots}.
The trained model is able to accurately predict \emph{Lot} peak stress from test images. Please refer to the supplementary material for details regarding the architecture, training, and performance of the model.
Despite the effectiveness of the predictive model, it is unclear how we can leverage the model for domain understanding and discovery, as the forward prediction process can only give us answers for existing material \emph{Lots} we have manufactured and scanned, which fails to provide actionable guidance in the synthesized process for producing material with more desirable attributes. In the next section, we will address such limitations by introducing a way to generate hypothetical \emph{Lots} for exploring the material design space.
}

\subsection{Attribute-Guided Image Generation}
\label{sec:imageSynthesis}
The generative adversarial network (GAN)~\cite{goodfellow2014generative} has revolutionized our ability to generate incredibly realistic samples from highly complex distributions~\cite{brock2018large, karras2019style}.
In general, a GAN transforms noise vectors ($\mathbf{Z}$ vectors from a high-dimensional latent space) into synthetic samples $\mathbf{I}$, resembling data in the training set.
The GAN is learned in an adversarial manner, in which a discriminator $D(\mathbf{I})$ (differentiate real vs. fake samples) and a generator $G(\mathbf{Z})$ (produce realistic fake samples) are trained together to compete with each other.
One limitation of the standard GAN is that the latent space is not immediately understandable, which limits our ability to control the generated content.
This problem is partially addressed by conditional GAN that is conditioned on the labels~\cite{mirza2014conditional}, i.e., generate different types of images by providing both a noise vector $\mathbf{Z}$ and a label $L$.
Still, these models, like most GANs, are often extremely hard to train and require a large number of samples for even moderately complex data.
Our initial attempts to apply conditional GAN on our SEM image data with the \emph{Lot} indices or other properties as labels were unsuccessful. This is likely due to an insufficient amount of images and labels as well as the innate complexity of the SEM image data.

To mitigate the training challenge and to improve the control over generated contents, we turn our focus to another class of GANs that makes selective modifications to existing images rather than generating them from scratch (i.e., transform a vector into the images).
Instead of providing a noise vector to the generator, these image editing GANs (e.g., attGAN~\cite{attGAN}) take an input image along with the attributes $\mathbf{A}$ that describe the desirable changes ($G(\mathbf{I}, \mathbf{A})$). For face images, such a GAN can be trained to alter attributes, such as the color of the hair or the presence of eyewear in the original image.
Since we provide the generator with an input image that already contains a large amount of information, we can build a model that not only produces a higher quality images but also requires fewer images to train than other classes of GANs.
More importantly, for our application, we can train such an attribute editing GAN, in which the material properties are the conditional attributes $\mathbf{A}$ that guides the image generation process. Such a model allows us to generate new images according to the given material attributes that are immediately understandable to the domain scientists. Next, we discuss the attributes used for modifying SEM images.

As discussed in Section~\ref{sec:background}, we have compressive strength measure for each \emph{Lot} of the material from laboratory tests.
To help understand material appearance in each \emph{Lot}, the material scientists estimated the following properties -- \emph{size}, \emph{porosity}, \emph{polydispersity}, \emph{facetness}, by examining a large number of images per-lot and averaging the estimates from multiple experts.
The estimated values are normalized (0 - 1). The meaning of each material property and specific features the scientists are looking for in the images are the following:
\emph{size} -- the average size of crystals;
\emph{porosity} -- how ``holey'' the crystals are, i.e., does it look like they have a lot of small pin-prick holes on the surface or are they solid;
\emph{polydispersity} -- how varied the size of the crystals are, i.e., how broad is the size distribution;
\emph{facetness} -- do the crystals look rounded/smooth at edges or do they have flat faces that meet at different angles to give a faceted structure.
As a result, only 30 labels/values per attribute are captured, which can be considered as extremely small for any traditional supervised learning task. Compared to other attributes in GAN applications, i.e., face images, in which we have individual defined labels for all images, the supervised \emph{Lot} level information for the SEM images are extremely sparse.


\SL{After obtaining these attributes, we train the attGAN~\cite{attGAN} that allows us to modify the material properties of a given SEM image (i.e., obtain an image from a hypothetical \emph{Lot} with the target material attributes).}
Besides the sparsity in labeling information, the other challenges originate from the presence of intricate patterns in the images itself.
For example, the porosity of a material is reflected by the presence of small pinprick holes on the surface of the crystals in the SEM image, which only occupies an extremely small number of pixels. Learning attributes represented by such a minuscule feature can be very challenging.
Despite these obstacles, as illustrated in Figure~\ref{fig:attEdit}, by utilizing the attGAN, the attribute-driven generation can accurately capture these intricate material features.
Such a success not only indicates the accuracy of the estimated material attributes by the scientists but also demonstrate the coherency among images from the same \emph{Lot}.

\SL{To ensure the image generation model is producing the intended modification, we examine the quality of generated SEM image from the following two aspects: 1) the synthesized images should be indistinguishable from the real SEM images, and 2) the generated images should exhibit material features that correspond to the modified attributes.
For a comprehensive analysis, we not only looked into widely adopted computational metrics but also investigated human perception through the feedback from material scientists. Both of these evaluations corroborated that the GAN-based SEM image editing process produces satisfactory results, i.e., meaningful images from a hypothetical \emph{Lot}.
To confirm the quality of the GAN model from a computational aspect, we closely examined the convergence and the loss behavior of both the generator, the discriminator, and the classifier in our model.
In particular, the low and stable reconstruction error indicates the GAN can reproduce realistic-looking SEM images. We also observed that the classifier can accurately predict the attributes from both the original and hypothetical (GAN-generated) images. This implies the generator can produce realistic modifications that can be correctly classified by the same classifier that correctly predicted attributes from the original images.
Moreover, we also resort to the material scientists for further evaluating the quality of generated images, as their domain knowledge is essential for understanding the intrinsic details and material concepts that may not easily be evaluated by the computational metrics.
According to the feedback from three material scientists, they not only had a hard time distinguishing between the images from original and hypothetical \emph{Lots} but also confirmed that the modification reflects the intended changes as described by the attribute inputs. These observations are also demonstrated in the examples of the attribute-guided modification as shown in Figure~\ref{fig:attEdit} (additional samples are also provided in the supplementary material).} We see in the top row, the larger crystal in the original image (left column) is naturally broken into smaller ones in the synthesized image that aim to decrease the overall \emph{size}.
Alternatively, we can see smaller crystals are removed (or suppressed) in the synthesized image to increase the overall crystal \emph{size} (highlighted by brown boxes).
In the second row, we can see that the small porous structures are being added in the rightmost image (increased \emph{porosity}), whereas the corresponding region is smoothed out in the middle image (decreased \emph{porosity}).
The \emph{polydispersity } attribute also works well, as we can see the GAN try to remove smaller crystal in the middle image (decreased \emph{polydispersity}) while increasing them in the case of increasing \emph{polydispersity}.
The \emph{facetness} is the only attribute that does not seem to be effectively isolated. Even though it appears to reduce/increase \emph{facetness} (see region marked by green and yellow squares), yet it also brings along more drastic change with respect to \emph{polydispersity} and \emph{size}. Moreover, there is likely an inherent dependency among these attributes, which we may not be able to eliminate even with additional data and labels.
For more examples, please refer to the supplementary materials.

\subsection{Actionable Explanation Pipeline}
\label{sec:forwardbackward}
As illustrated in Figure~\ref{fig:pipeline}, once an image from the hypothetical \emph{Lot} is generated, we can feed it into the predictive model to predict the respective mechanical properties (e.g., peak stress).
As discussed in Section~\ref{sec:background}, one of the goals for building the regression model is to better understand the relationship between SEM images and the mechanical properties of the respective \emph{Lots}.
The introduction of the attribute-driven image generation process not only exposes the explicitly defined material features but also enables the ability to actively control them to form intervention operations that are essential for reasoning about counterfactual relationships (i.e., alter a material feature and then observe corresponding changes in the prediction).
An added benefit of the image editing GAN is that it often strives to introduce minimal alteration in the image for required attribute change (e.g., for face image, the attGAN can change the hair color without altering other facial features).
Such behavior makes it suitable to reason about the effect of the change, as the editing does not intend to change other features or the general structure of the original image.

The most straightforward way to ascertain the relationship between the material attributes and the predicted mechanical properties is to do a simple ``forward'' sensitivity analysis by observing how predicted stress changes as we vary the material properties in the image generation process.
To understand the impact of a particular set of attributes, we can fix all other attributes while varying the values of the attributes of interests. We then feed the generated images to the predictive model and obtain the corresponding predicted peak stress (see Section~\ref{sec:result} for more details).
\SL{
Such an analysis allows us to estimate the sensitivity (i.e., importance) for each of the material attributes prediction, which enables material scientists to form intuition about the influence of the attribute changes on the peak stress of a given \emph{Lot}.
}


\BK{However, in several scenarios, we may be interested in answering retrospective questions that can provide precise actionable insights to improve the performance of a certain \emph{Lot}, e.g., what specific changes should be made to the attributes of a given \emph{Lot} to increase its peak stress?}
To address this challenge, we introduce a ``backward'' explanation scheme, in which an optimization is performed to obtain the necessary changes to the input attributes for obtaining the desired peak stress change.
Let us define the generative editing model as ${G}(\mathbf{I};\mathbf{A})$, where $\mathbf{I}$ is the original image and ${\mathbf{A}}=\{a_1,\cdots,a_N\}$ are the material attributes that control the editing.
Given an SEM image $\mathbf{I}$ for which the regressor $R$ predicts a certain peak stress, we aim to identify attributes $\mathbf{A'}$ with minimal deviation from $\mathbf{A}$ such that the edited image $\mathbf{I(A')}={G}(\mathbf{I};\mathbf{A'})$ would lead to a higher/lower peak stress prediction $p$.
Given an image $\mathbf{I}$ with corresponding image attribute vector $\mathbf{A}$ and a target peak stress $p$, we formulate backward explanation problem as follows: 

\begin{equation}
\label{opt}
\begin{aligned}
\min_{\mathbf{A}'} \quad & \|\mathbf{A}-\mathbf{A'}\|_q\\
\textrm{s.t.} \quad & p=R(\mathbf{I}(\mathbf{A}'))\\
  & \mathbf{I(A')} = G(\mathbf{I};\mathbf{A}').    \\
\end{aligned}
\end{equation}

The neural network model makes the formulation \eqref{opt} non-linear and non-convex that makes it difficult to solve the problem in its original form. Thus, we formulate a relaxed version of the problem that can be solved efficiently as follows

\begin{equation}
\label{opt1}
\begin{aligned}
\min_{\mathbf{A'} }\quad & \lambda \cdot \|p - R(G(\mathbf{I};\mathbf{A}'))\|_2 +  \|\mathbf{A} - \mathbf{A}')\|_1,\\
\end{aligned}
\end{equation}

\SL{where mean squared error (MSE) loss is used to encourage the predicted peak stress to be closer to the target peak stress $p$. Further, to obtain a more sparse (i.e., understandable) explanation, we set the $q=1$ in the regularization term.}
Since, both regressor $R$ and generator $G$ are differentiable, we can compute the gradient of the objective function via back-propagation, and solve the optimization problem using gradient descent algorithm.

\SL{
The backward explanation enables us to answer the retrospective questions by automatically identifying specific attribute changes that can lead to the target peak stress. Such an analysis not only facilitates a direct way for examining the effects of simultaneous modification of multiple attributes but also produces the actionable guidance for the synthesis process for achieving a more desirable material property.}
Despite the model's aim to predict peak stress of the given \emph{Lot}, the prediction itself is made based on a single image tile (each \emph{Lot} contains a large number of image tiles, see details in Section~\ref{sec:background}). As a result, it is imperative to look beyond the behavior of individual prediction and examine the average behavior of all image tiles of a specific \emph{Lot}. The same applies to the model explanation, in which we can obtain a more comprehensive understanding of the behavior of the \emph{Lot} by averaging or aggregating its explanation (e.g., utilizing boxplot, see Section~\ref{sec:result} for details).

%% file: results.tex
\section{Results and Discussions}
\label{sec:result}

Here we illustrate the application of the proposed techniques to help material scientists obtain actionable insights from the regression model and infer the underlying relationship between feedstock materials’ characteristics and their compacted mechanical performance.


\begin{figure*}[!t]
\centering
  \includegraphics[width=0.99\linewidth]{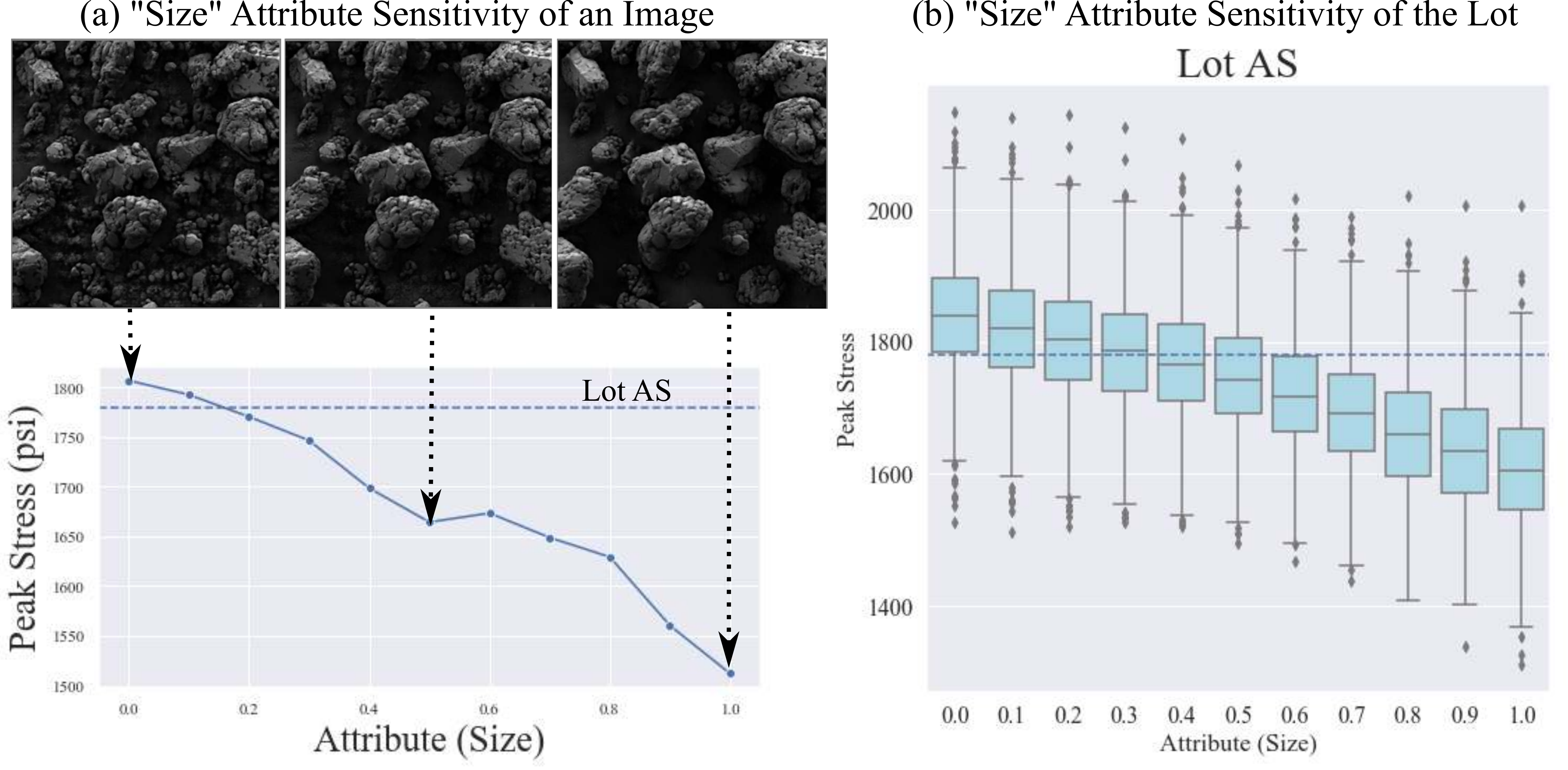}
 \caption{
\SL{
Illustration of the forward explanation. In (a), the effect of varying the attribute of a single input image tile is illustrated. Here we explore the images from different hypothetical \emph{Lots} by varying the crystal size.
The x-axis is the normalized material attribute values (0-1), which corresponds to the relative strength of the changes, e.g., zero indicates a very small size with respect to the norm in the given \emph{Lot}.
The horizontal dotted line illustrates the measured peak stress of the given \emph{Lot}.
We can also perform similar aggregated analysis on all image tiles from a specific \emph{Lot} (in this case \emph{Lot} AS) through a boxplot~\cite{potter2006methods} as illustrated in (b).}}
\label{fig:forwardExplainDemo}
\end{figure*}

The proposed technique allows the integration of material attributes (such as, crystal \emph{size}, \emph{porosity}, \emph{polydispersity}) as tunable knobs in the analysis pipeline. As a result, material scientists can intuitively reason about the impact of varying a given attribute on the predicted peak stress.
As illustrated in Figure~\ref{fig:forwardExplainDemo}(a), by altering the \emph{size} attribute when generating the images of hypothetical \emph{Lots}, we can observe changes in the predicted peak stress values. Here, we generated $11$ images with \emph{size} attribute varying from 0.0 to 1.0 (the full range of the attribute) while fixing all other attributes. Attribute values are shown in the x-axis, whereas the predict peak stress (in \emph{psi}) is shown in the y-axis. As shown in Figure~\ref{fig:forwardExplainDemo}(a), for a single input SEM image instance, the predicted peak stress decreases as we increase the crystal particle size.
The visual effects of the size attribute change can also be observed in the corresponding images (only three images are shown due to space constraint).
Since our regression model generates peak stress prediction for a given \emph{Lot} based on a single image tile (each \emph{Lot} image contains thousands of image tiles), certain variation exists among the image tiles within each \emph{Lot}. Therefore, for evaluating the model behavior it is crucial to understand the average behavior of predictions for all image tiles.
As shown in Figure~\ref{fig:forwardExplainDemo}(b), we show the aggregated results from all tiles from the \emph{Lot AS}.
In the boxplot, each vertical glyph (along the x-axis that corresponds to attribute value size) encodes predictions of all image tiles with the same attribute values. The y-axis shows the predicted peak stress.
Despite the variation among the tiles in the \emph{Lot}, we can observe a similar trend in both (a) and (b).

\begin{figure*}[!t]
\centering
  \includegraphics[width=0.99\linewidth]{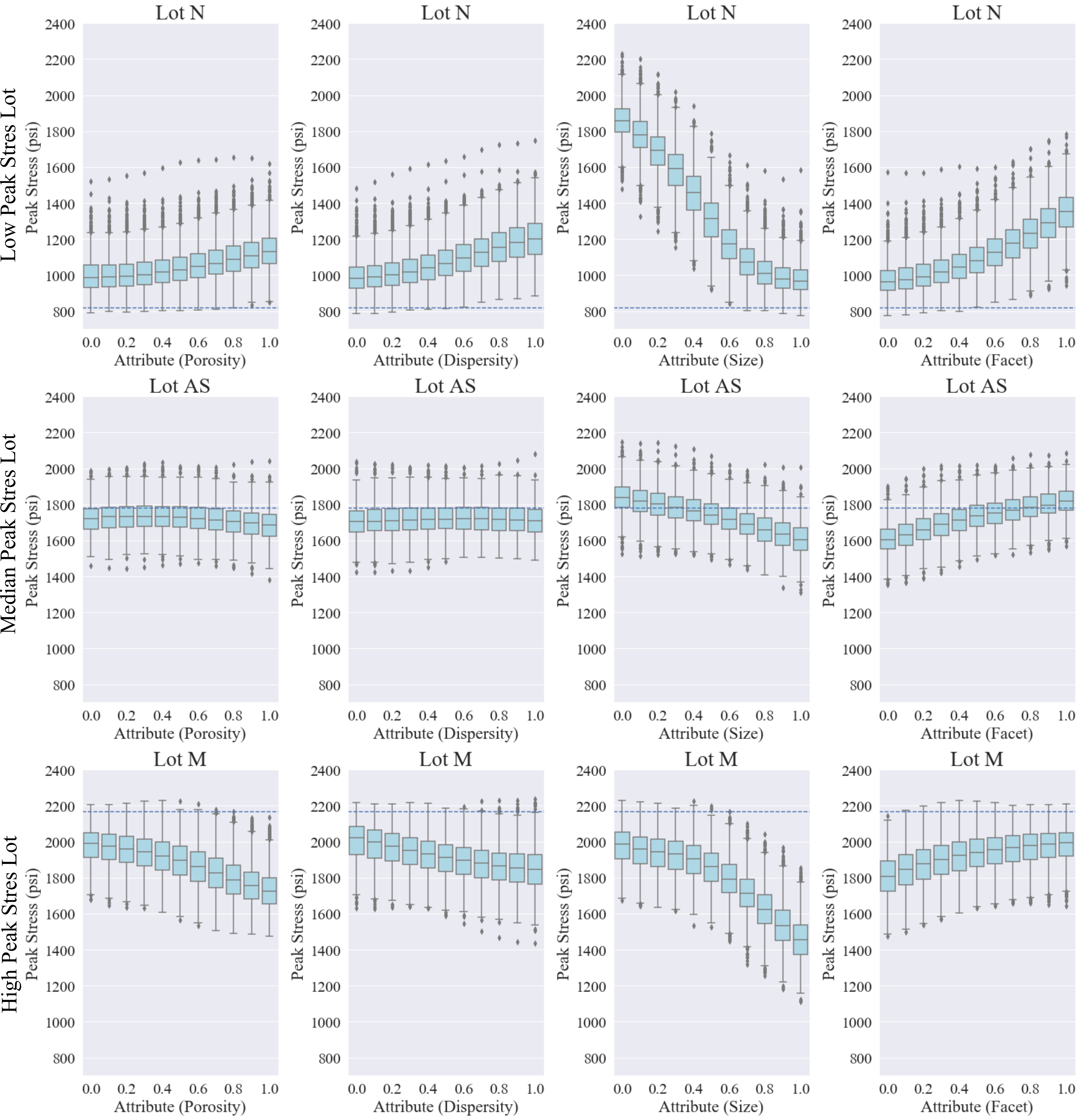}
 \caption{
  Forward explanations illustrate the sensitivity of the peak stress with respect to varying attribute values for different \emph{Lots}.
}
\label{fig:forwardExplainLot}
\end{figure*}

Similarly, we can evaluate how each of the four attributes impacts the peak stress prediction by applying a similar sensitivity analysis for different \emph{Lots} by varying the attribute values one at a time.
In Figure~\ref{fig:forwardExplainLot}, we illustrate three \emph{Lots}, with high (M), median (AS), and low (N) peak stress values, respectively. As shown in the plots, the \emph{size} and \emph{facet} attributes have a pronounced and consistent effect on the prediction output, which shows that {\em having larger particles in general has a detrimental impact to the compressive peak performance of the sample, while having more well faceted crystals in the samples are beneficial to increasing the peak stress values}.
Our analysis also highlights that there are no clear trends for both \emph{porosity} and \emph{polydispersity} attributes, which diverges depending on the selection of the \emph{Lot}. The divergence in these attributes show that {\em there is more than one single pathway to achieve a particular peak stress value}, since the exemplar cases of M, AS, and N \emph{Lots} all have very different original attributes.  The ability to edit and modify attributes from a distinct original point to either increase or decrease the desired performance provides powerful visualization cues to the subject matter experts while also informing which knobs should be tuned (and their sensitivities) to achieve the desired performance.
%


\begin{figure*}[htbp]
\centering
  \includegraphics[width=0.99\linewidth]{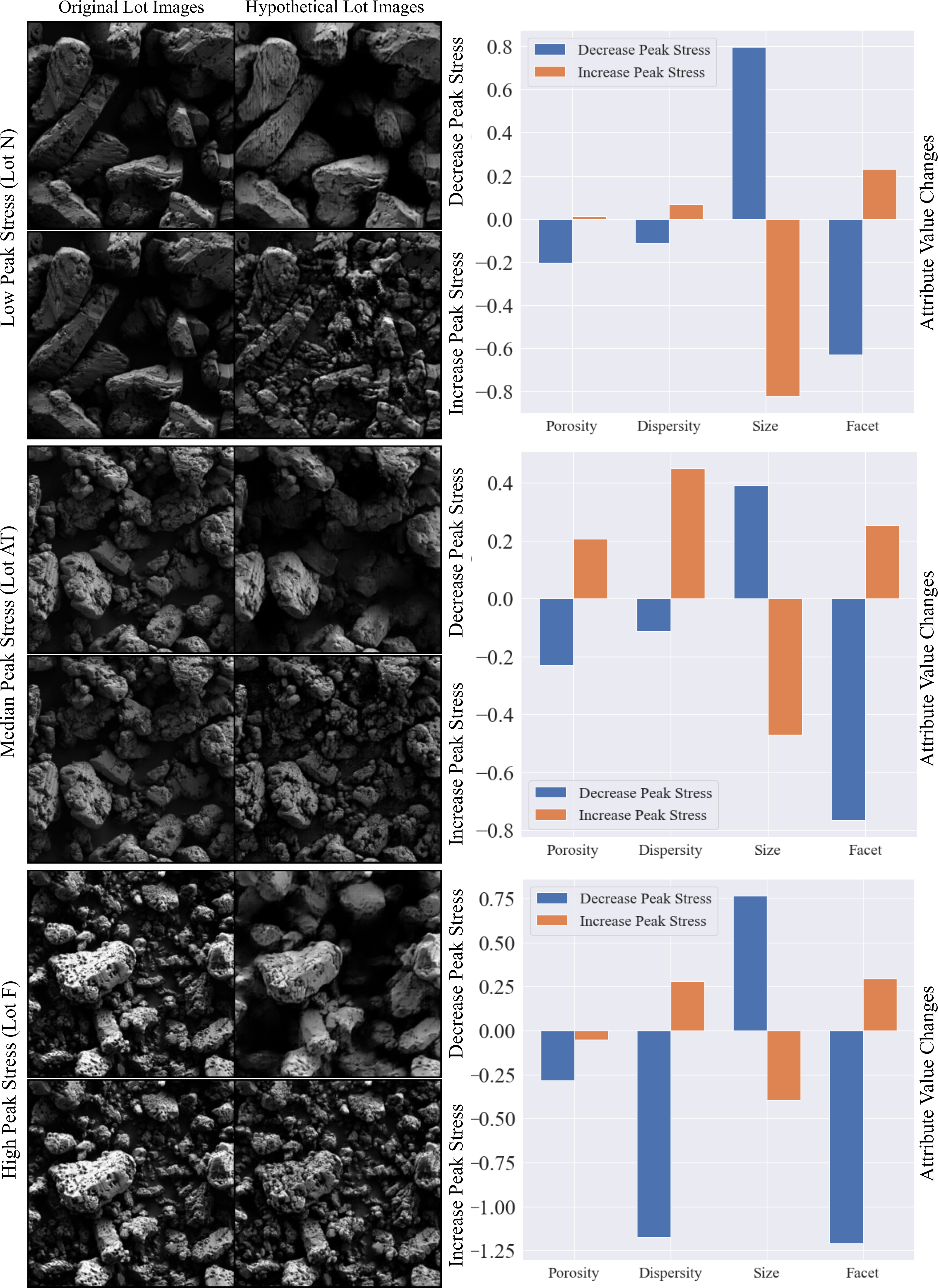}
 \caption{
 Backward actionable explanation for a single SEM image. The images from original and hypothetical \emph{Lot} (through GAN-based image manipulation) are shown on the left, and the corresponding attribute changes that led to an increase or decrease of the predicted peak stress are illustrated in the bar chart on the right. 
}
\label{fig:backwardPerInstance}
\end{figure*}

As discussed in Section~\ref{sec:forwardbackward}, the sensitivity analysis that utilizes forward evaluation only allows us to discern the impact of varying one (or limited combination) attribute. To truly understand the interaction among them, we introduced backward explanation, in which we ask what are the minimal changes that can be made to the attribute so that the model would alter the prediction to a desirable value.
%
In Figure~\ref{fig:backwardPerInstance}, the original image and modified image for increasing and decreasing predicted peak stress (based on the changes in the attributes) are shown.
In the top row (Lot N), we can see in both SEM images (left) and attribute bar-plot (right), that decreasing crystal \emph{size}, while increasing \emph{porosity}, \emph{polydispersity}, \emph{facetness} will lead to higher peak stress prediction.
The same pattern can be observed for \emph{Lot} AT (mid-row).
Although bottom row (Lot F) shows a slight deviation compared to earlier patterns on porosity, the small absolute value indicates the change in \emph{porosity} does not contribute much to the changes in the generated image.
One thing to note is that the increase of the \emph{facetness} attributes in the image generation process seems to also lead to a marked increase in \emph{polydispersity} and a reduction of average \emph{size} (see Section~\ref{sec:imageSynthesis}), so the effect we observe for altering \emph{facetness} is likely also due to the changes in \emph{size} attribute.

\begin{figure*}[htbp]
\centering
  \includegraphics[width=0.99\linewidth]{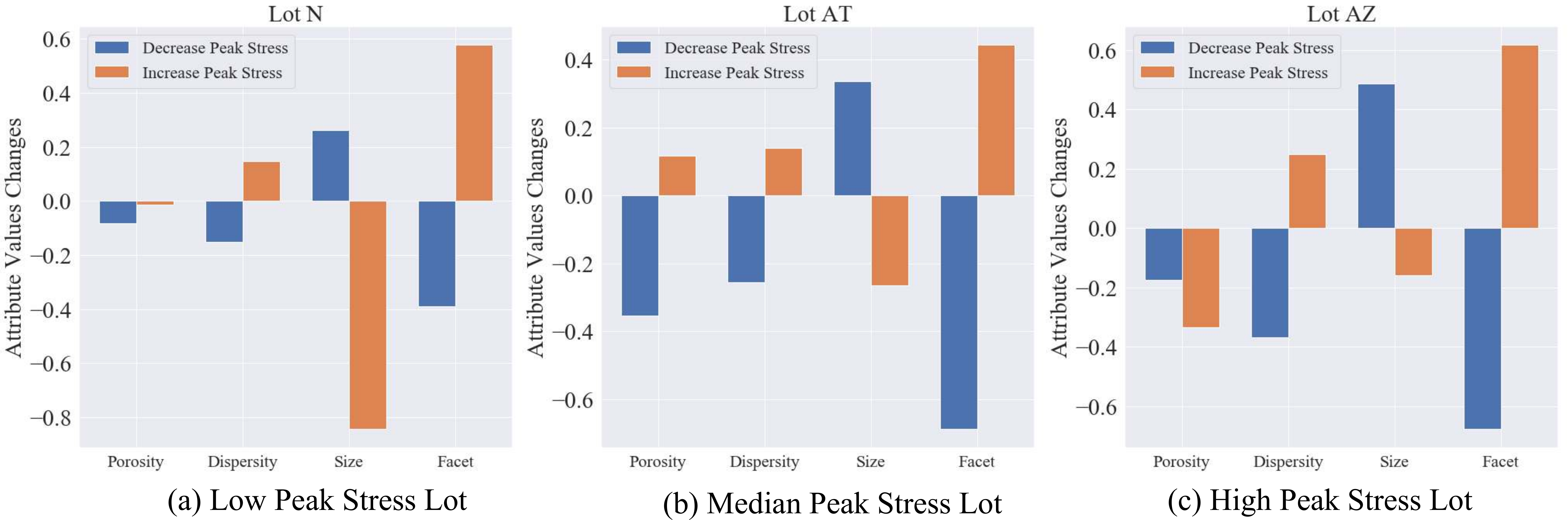}
 \caption{
  Backward explain result for the given \emph{Lots} by averaging per-image explanation.
}
\label{fig:backwardPerLot}
\end{figure*}

We also estimate the overall behavior of the entire \emph{Lot} by averaging the backward explanation of all image titles from a given \emph{Lot}.
As shown in Figure~\ref{fig:backwardPerLot}, we can utilize a similar attribute change plot to illustrate the averaging behaviors by showing the mean values.
The plots show that the rule we identified by examining backward explanation of individual images is consistent with average behavior of the entire \emph{Lot}.
The \emph{polydispersity}, \emph{size}, and \emph{facetness} behave consistently in increasing/decreasing of the peak stress.
Since both \emph{polydispersity} and \emph{size} attributes are directly associated with the mean and variance of crystal size, \emph{the reduction of size appears to be the most effective route to increase peak stress prediction}.  

In discussion with subject matter experts (SMEs), grinding of TATB is a common practice to increase peak strength for samples which does not meet the desired peak stress requirement. According to the SMEs, the grinding process increases the number of smaller crystal sizes, (thereby lowering the overall crystal sizes) while increasing the polydispersity of the samples. Increasing polydispersity in particle distributions is a common approach to increasing particle packing densities~\cite{sohn1968effect} in a number of different applications.

\SL{
Compared to the conventional wisdom, our method not only explicitly confirms the impact of the crystal sizes, but also produces realistic depiction of the appearance of the material for which the given predictive model would predict to have a higher peak stress value.
Moreover, our method enables a multifaceted analysis from a single instance of the prediction to the averaging behavior of the entire \emph{Lot}, from the sensitivity of a single attribute to the joint influence of multiple ones for achieving an optimal objective value.
Our approach will also be valuable in emerging applications where scientists have not yet formed deep scientific insights and can serve as a useful computation tool by providing actionable explanations without extensive experiments.}

%% file: discussion.tex
\section{Conclusion}
We introduced a general technique for inferring actionable insights from a given predictive model by understanding and manipulating the domain attributes. The ability to turn these explainable ``knobs'' allows us to obtain counterfactual understanding on how the prediction is affected by key domain attributes.
To better understand the combined effects of multiple attributes, we introduced an optimization algorithm that allow the model to help reveal what attribute combinations would yield a more desirable output.
For domain scientists to adopt the emerging machine learning techniques, domain specific explanations of how the machine learning models are functioning is essential.  Without tangible and actionable information from machine learning models, the overall benefit machine learning will have in scientific domains is limited. The work presented here demonstrate that it is possible gain actionable insights from complex machine learning pipelines that can accelerate the materials development processes.  It is also important to note that our ability to meaningfully modify and generate hypothetical SEM images based on domain attributes is driven by the recent development in image editing GANs~\cite{attGAN}.
\SL{
Moreover, since we obtained the explanation through controlling the attribute-aware variation in the input data, compared to many state-of-the-art explanation techniques, the proposed technique is not restricted to a specific model and can be adapted to understand the behavior of other predictive models as well.}
As with any newly developed techniques, our approach has some shortcomings that needs future improvement. One particular challenge originates from the potential distribution shift from original image to the reconstructed images (when we generate new image tiles using the attributes associate the corresponding \emph{Lot}). Even though a human viewer often cannot discern any noticeable difference between the original images and reconstructed ones, these unnoticeable changes can lead to minor prediction shifts from the original ones.
Moreover, due to the inherent limitation of how the regression model is built, we try to predict the peak stress for the given \emph{Lot} based on a single SEM image tile and average the predictions, which leads to built-in variation among predicted values generated from different image tiles from the same \emph{Lot}.  We are currently exploring other approaches to build more robust regression models that capture the overall qualities of the samples from limited data (i.e, data efficient model design~\cite{mallick2019deep}), a common obstacle in applying machine learning to scientific data.

%% file: supplemental.tex
\appendix
\section*{Supplemental Material}

\section{SEM Image Dataset}
The SEM dataset we used for training and evaluating both the regressor and generative model consists of 59,690 greyscale images, with a resolution of $256 \times 256$ (downsampled from $1000 \times 1000$ in the original image). These images are from 30 unique \emph{Lots} (batch of material), in which compressive strength testing is carried out for each \emph{Lot} to obtain the corresponding peak stress value.

\section{Detail of the Peak Stress Prediction Model}

\textbf{Model architecture:} The regression model architecture for peak stress prediction is based on the Wide ResNet model \cite{zagoruyko2016wide}, with a total of 28 convolutional layers and a widening factor of 1, followed by an adaptive average pooling layer. Since the Wide ResNet model was originally proposed for classification, we also need to replace the final \emph{softmax} layer with a fully connected regression layer \emph{tanh} activations to predict continuous scalar values. Our implementation is based on PyTorch.

\noindent\textbf{Training setup:}  We set aside 10\% of the training data for validation, leaving a total of 53721 training images and 5969 validation images. All images are preprocessed by subtracting the mean and dividing by the standard deviation. For data augmentation, we do horizontal flips. We train the regression model with the mean squared error (MSE) loss function and the Adam optimizer \cite{kingma2014adam} with a learning rate of 0.001 and a minibatch size of 64. We used early stopping to terminate training when the validation performance did not improve, and the whole training procedure stops in 48 epochs.

\begin{figure*}[htbp]
\centering
  \includegraphics[width=0.8\linewidth]{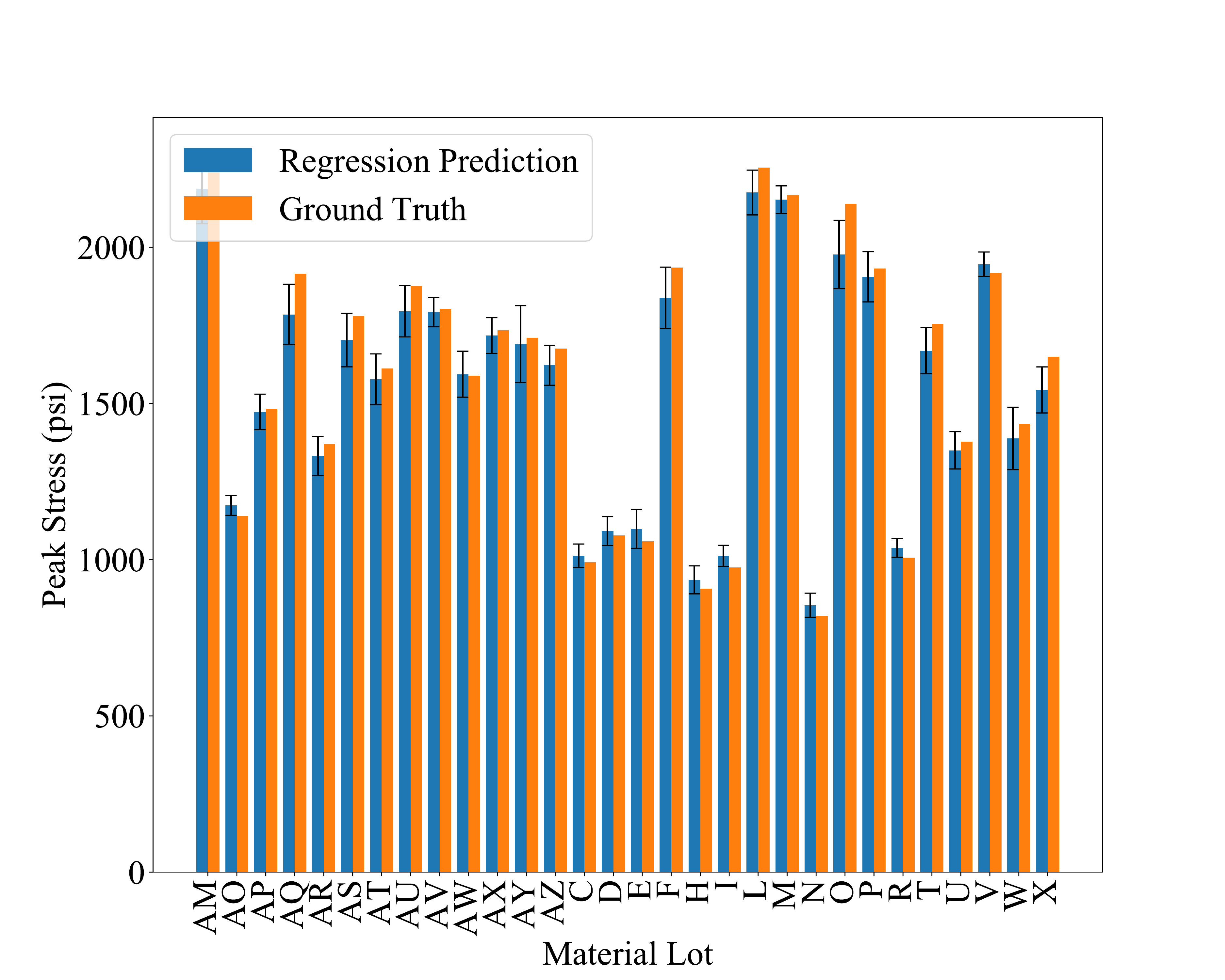}
 \caption{
  The predicted and ground-truth value of peak stress for different \emph{Lots}.
}
\label{fig:predictionPerLot}
\end{figure*}

\begin{figure*}[htbp]
\centering
  \includegraphics[width=0.8\linewidth]{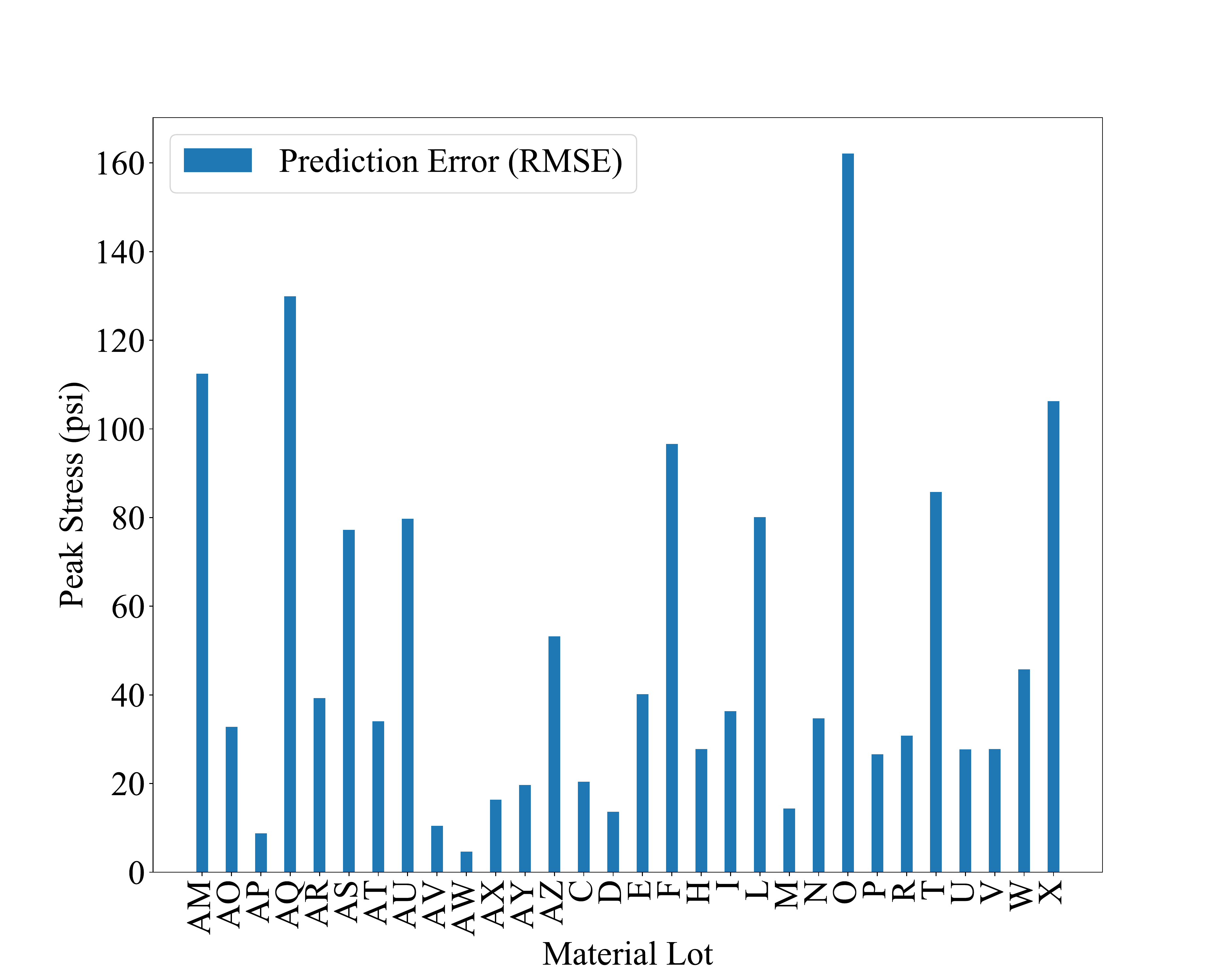}
 \caption{
  RMSE of the predicted peak stress values for different \emph{Lots}.
}
\label{fig:rmsePerLot}
\end{figure*}

\noindent\textbf{Prediction performance:} Globally, the regression model achieved a root mean square error (RMSE) of $66.0$ and a mean absolute percentage error (MAPE) of 3.07\% across all \emph{Lots}. For each \emph{Lot}, the peak stress predictions versus the ground-truth peak stress values are shown in Figure~\ref{fig:predictionPerLot}, where the error bars present the standard deviation of predictions across images in the \emph{Lot}.
The root mean square error per \emph{Lot} is plotted in Figure~\ref{fig:rmsePerLot}.

\section{Training and Evaluation of Image Editing GAN}
\noindent\textbf{Training setup:} As discussed in Section~\ref{sec:method}, our result for SEM image is achieved by training the  AttGAN~\cite{attGAN} with the material attribute labels provide by material sciences.
Compared to training setup for celebrity image dataset, the largest different with SEM images is the number of available labels.
For the celebrity images, labels are obtained for each individual face images. However, it is the case for the SEM images, in which we only have label for each \emph{Lot} that contain large number of images.
We trained the attGAN utilizing a pytorch implementation  with the same learning rate as for the celebrity dataset.
The output image of the GAN is same as input with a resolution of $256 \times 256$.
We trained the model for 70 epochs. The training cutoff is determined by examining the sample results during training, where additional epochs do not appear to improve the visual fidelity of the generated modification.

\begin{figure*}[htbp]
\centering
  \includegraphics[width=0.99\linewidth]{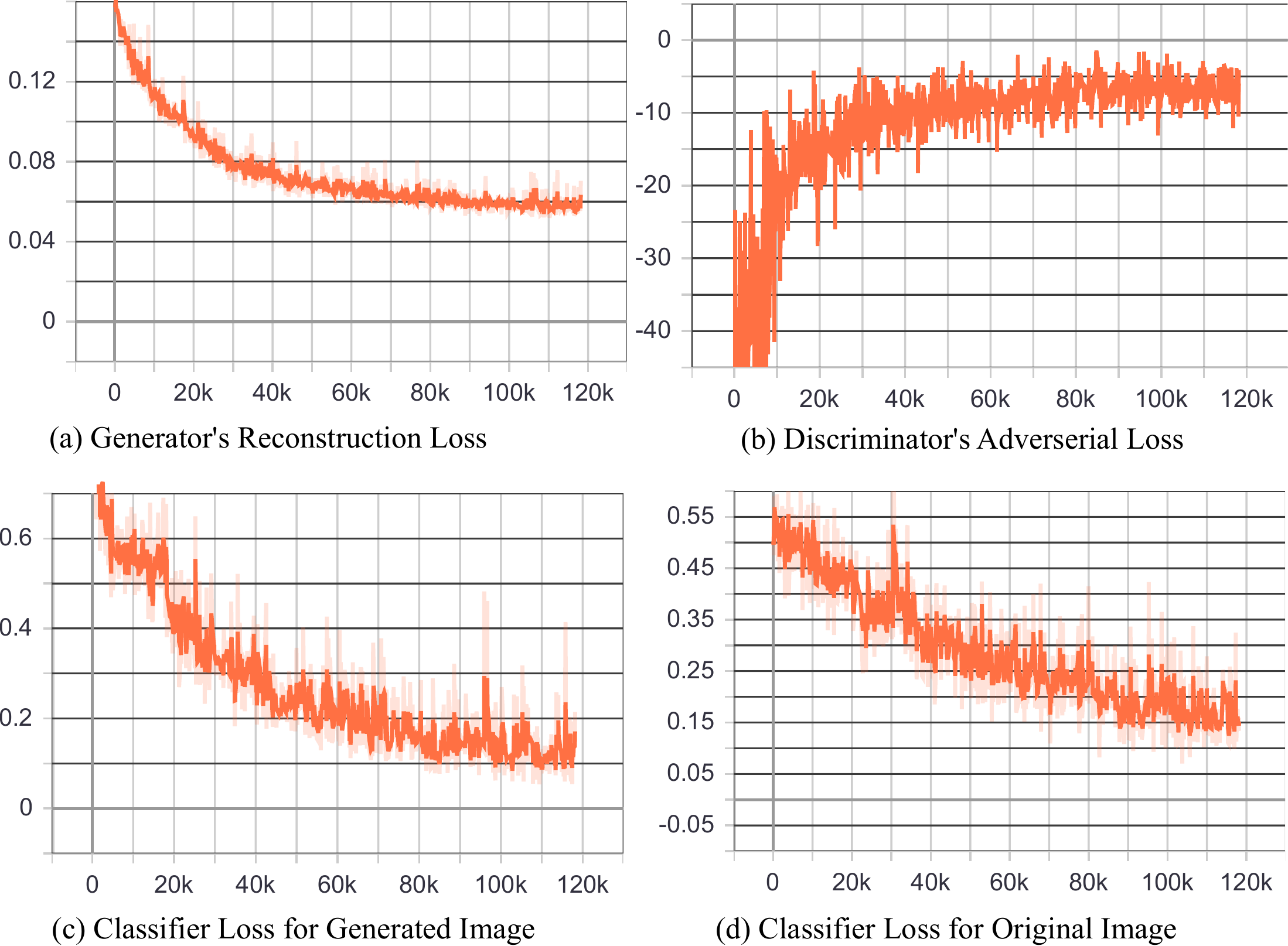}
 \caption{
  The performance curves for the GAN training. (a) illustrates the generator's reconstruction loss. As the error decreases, the generator is able to produce more realistic SEM images. (b) shows discriminator's adversarial loss, which increases over training iterations as the generator is producing more realistic images, in turn, fooling the discriminator.
  The (c) and (d) show the classification loss for predicting the attribute labels for both original (part of the overall training loss for the discriminator) and generated (part of the overall training loss for the generator) SEM image.
}
\label{fig:GAN_training}
\end{figure*}

\noindent\textbf{Training Evaluation:}
The training details for the GAN is shown in Figure~\ref{fig:GAN_training}.
The attGAN jointly trains the generator (contains both encoder and decoder), the discriminator, and the classifier for predicting attributes. Please refer to the original work~\cite{attGAN} for more details about model architecture.
As shown in Figure~\ref{fig:GAN_training}, (a) illustrates the generator's reconstruction error. As the error decreases, the generator is able to produce more realistic SEM images. (b) shows discriminator's adversarial loss. The loss increases over training iterations denoting that the generator is producing more realistic images, in turn, fooling the discriminator.
The (c) and (d) show the classification loss for predicting the attribute labels from both original (part of the overall training loss for the discriminator) and generated (part of the overall training loss for the generator) SEM images.
The decreasing and then stabilizing behavior of the classifier losses indicates the jointly trained classified can accurately predict the attributes for both the real and fake (generated) images. This also implies that the generator can produce realistic modifications that can be correctly classified by the same classifier that is predicting attributes from the real images correctly.
Moreover, the low and stable reconstruction indicates the generator can reproduce realistic-looking SEM images.
These combined observations provide the evidence to support our claim that the GAN is trained well and the quality of the image editing process is good as showcased by the classifier performance on generated images.

\section{Additional Examples for the GAN Modified SEM Images}
\begin{figure*}[!htbp]
\centering
  \includegraphics[width=1.0\linewidth]{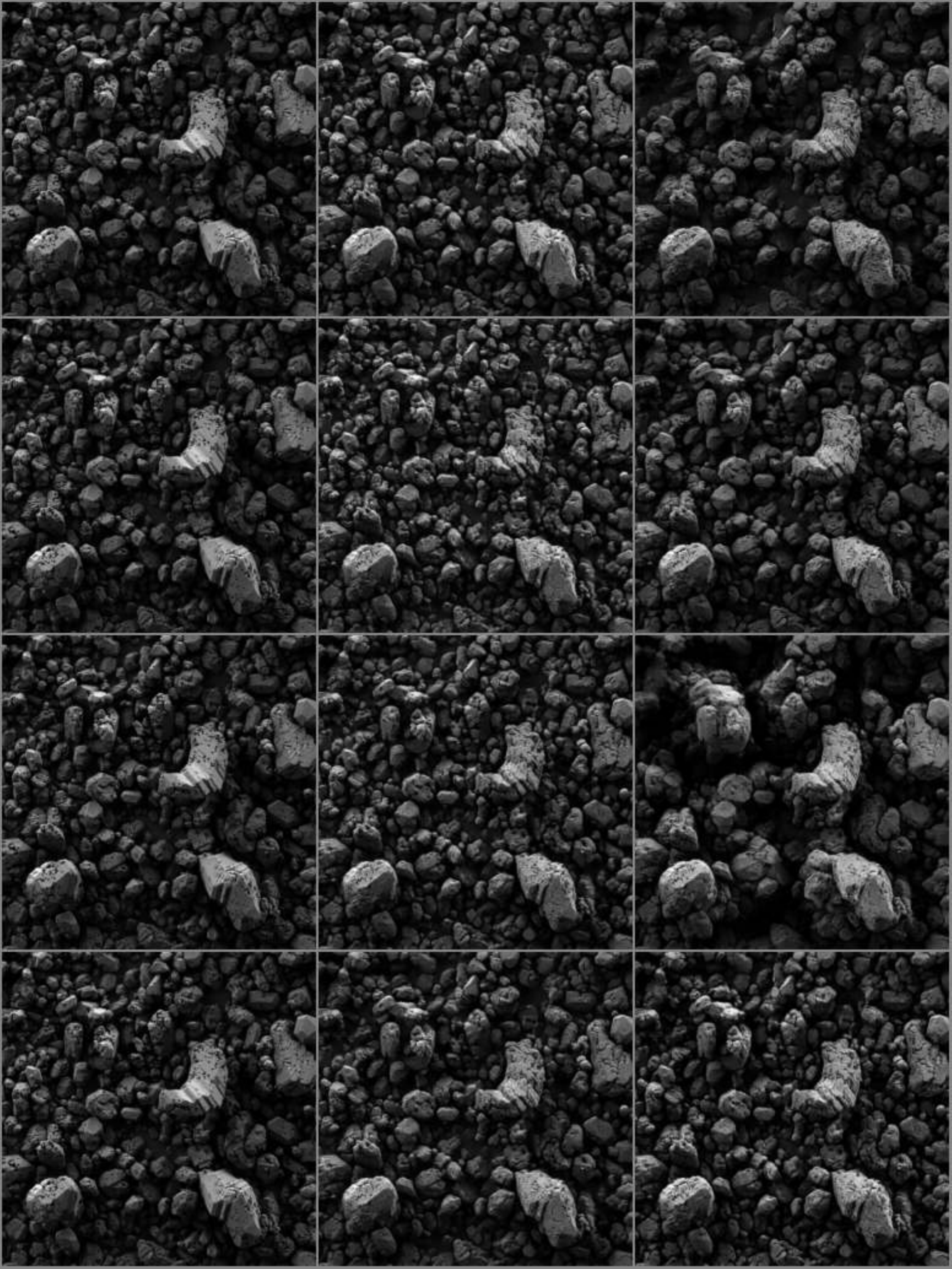}
 \caption{
    Additional examples of GAN generated images. Similar to the corresponding figure in the main text, the left column is the original SEM image. The middle column images are synthesized to increase the corresponding attributes, whereas the right column is synthesized to decrease the corresponding attribute. The four rows correspond to the four attributes, namely, \emph{porosity}, \emph{polydispersity}, \emph{size}, and \emph{facetness}
}
\label{fig:additionalResults_1}
\end{figure*}

\begin{figure*}[!htbp]
\centering
  \includegraphics[width=1.0\linewidth]{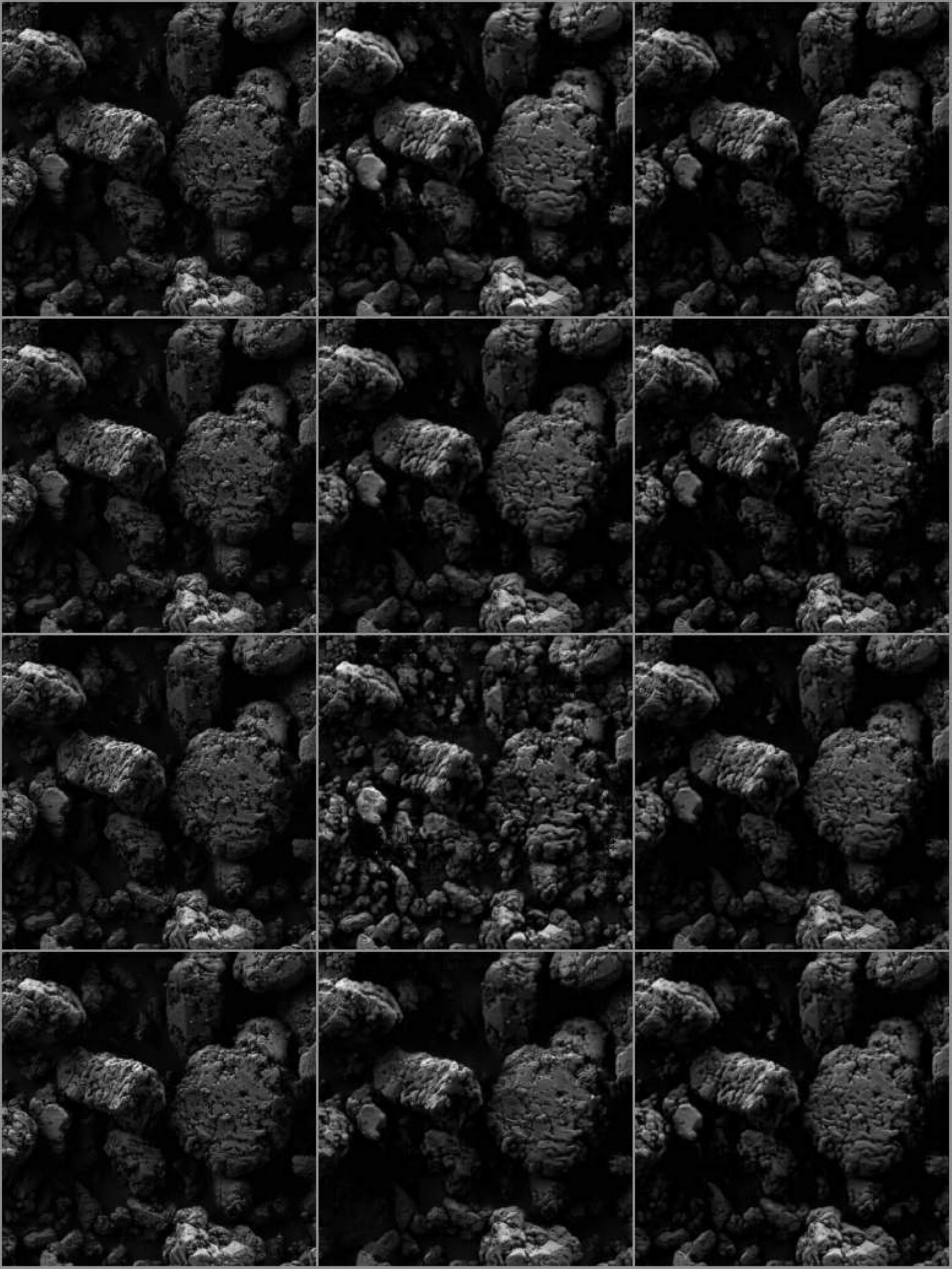}
 \caption{
 Additional examples of GAN generated images. Similar to the corresponding figure in the main text, the left column is the original SEM image. The middle column images are synthesized to increase the corresponding attributes, whereas the right column is synthesized to decrease the corresponding attribute. The four rows correspond to the four attributes, namely, \emph{porosity}, \emph{polydispersity}, \emph{size}, and \emph{facetness}
  }
\label{fig:additionalResults_2}
\end{figure*}

\begin{figure*}[!htbp]
\centering
  \includegraphics[width=1.0\linewidth]{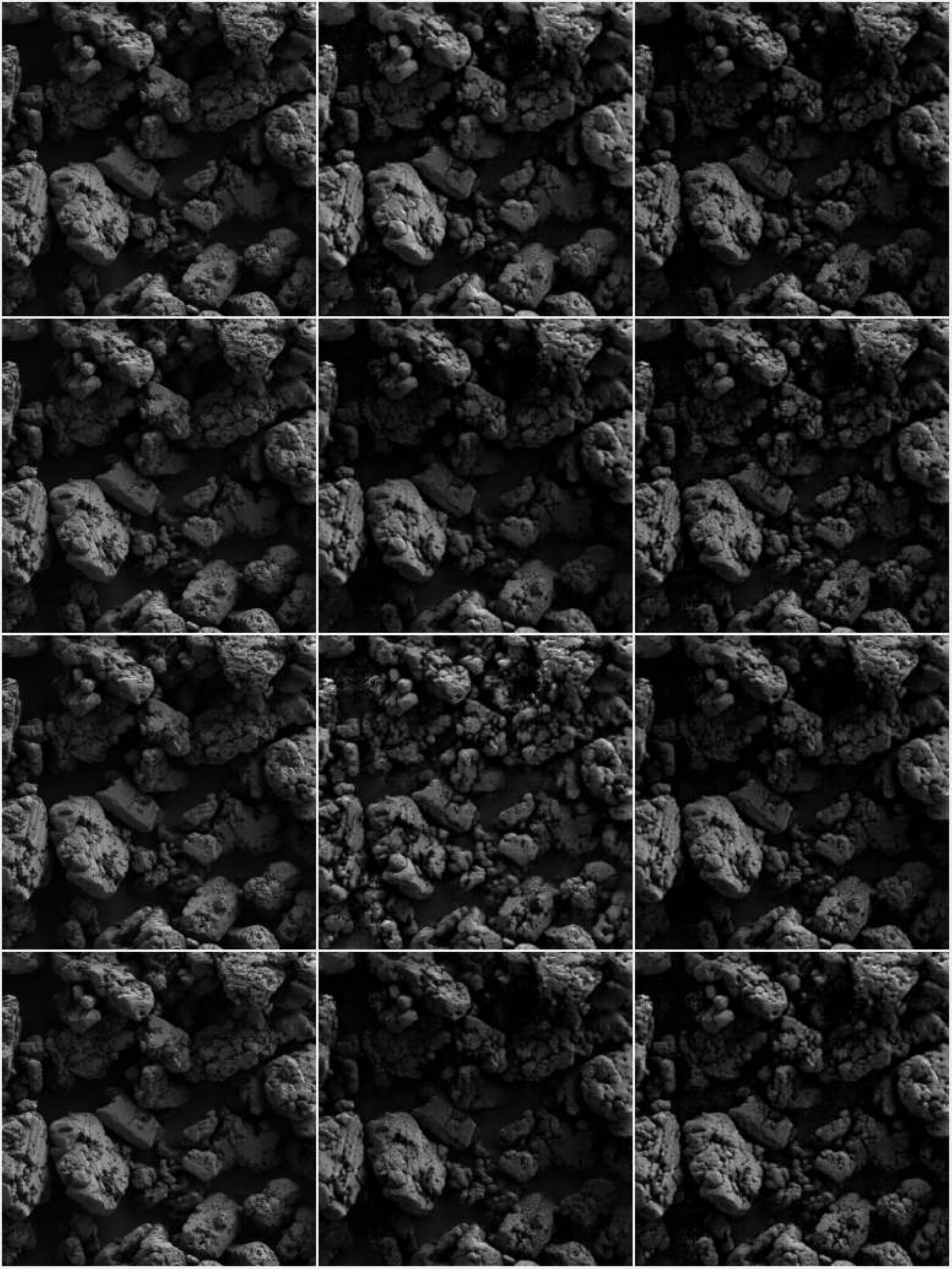}
 \caption{
  Additional examples of GAN generated images. Similar to the corresponding figure in the main text, the left column is the original SEM image. The middle column images are synthesized to increase the corresponding attributes, whereas the right column is synthesized to decrease the corresponding attribute. The four rows correspond to the four attributes, namely, \emph{porosity}, \emph{polydispersity}, \emph{size}, and \emph{facetness}}
\label{fig:additionalResults_3}
\end{figure*}

\begin{figure*}[!htbp]
\centering
  \includegraphics[width=1.0\linewidth]{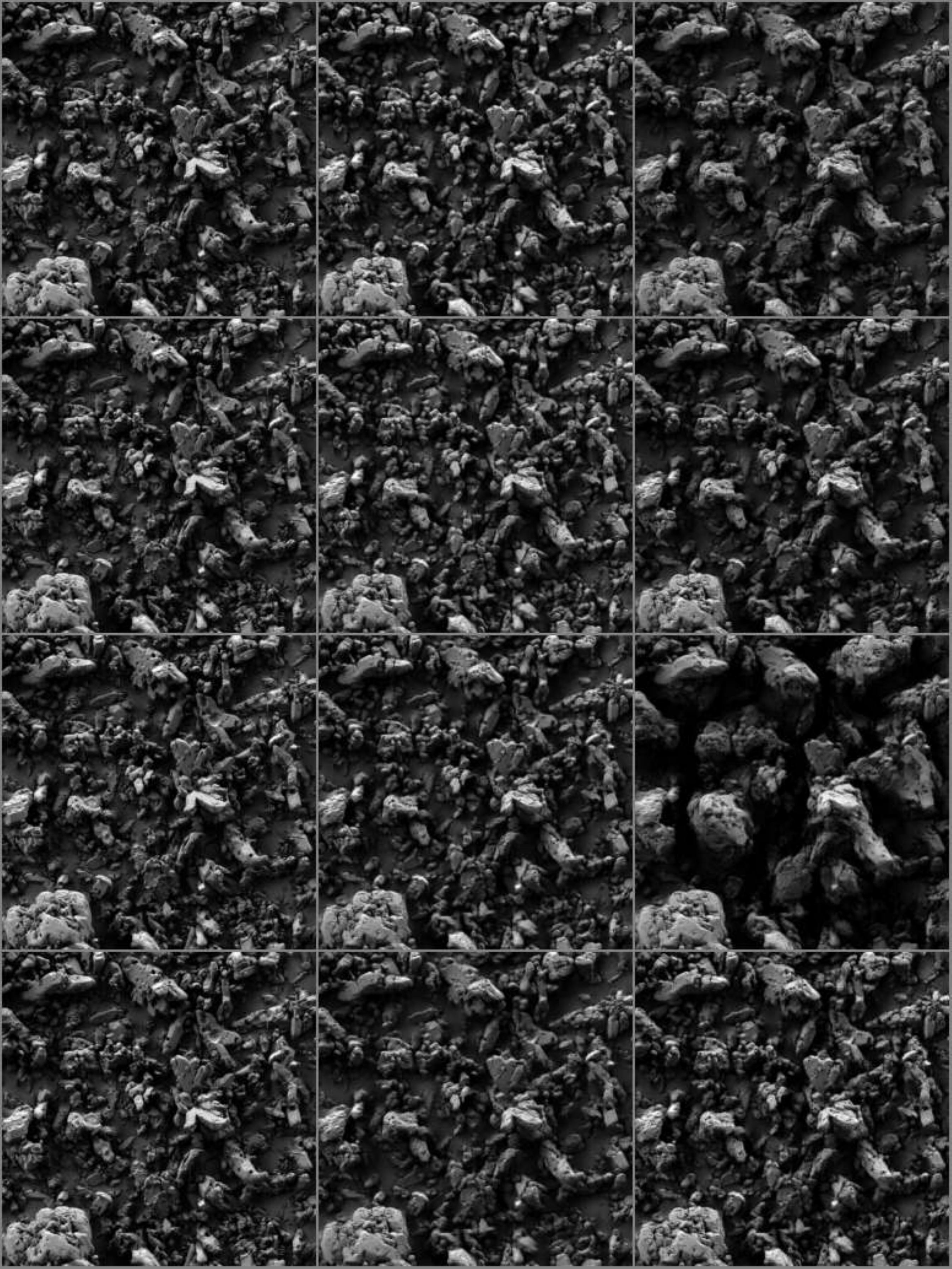}
 \caption{
  Additional examples of GAN generated images. Similar to the corresponding figure in the main text, the left column is the original SEM image. The middle column images are synthesized to increase the corresponding attributes, whereas the right column is synthesized to decrease the corresponding attribute. The four rows correspond to the four attributes, namely, \emph{porosity}, \emph{polydispersity}, \emph{size}, and \emph{facetness}}
\label{fig:additionalResults_4}
\end{figure*}

\begin{figure*}[!htbp]
\centering
  \includegraphics[width=1.0\linewidth]{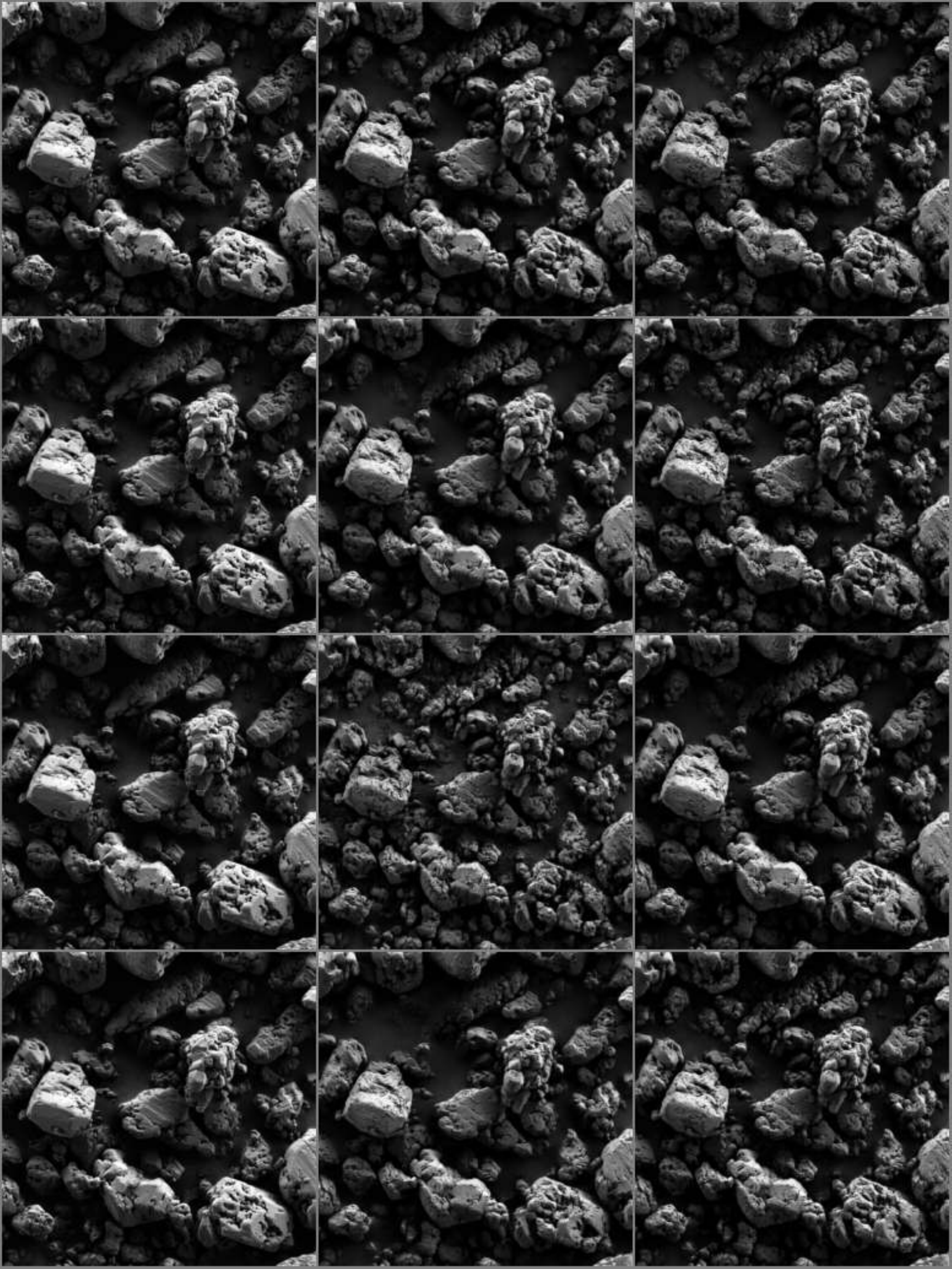}
 \caption{
  Additional examples of GAN generated images. Similar to the corresponding figure in the main text, the left column is the original SEM image. The middle column images are synthesized to increase the corresponding attributes, whereas the right column is synthesized to decrease the corresponding attribute. The four rows correspond to the four attributes, namely, \emph{porosity}, \emph{polydispersity}, \emph{size}, and \emph{facetness}}
\label{fig:additionalResults_5}
\end{figure*}

\begin{figure*}[!htbp]
\centering
  \includegraphics[width=1.0\linewidth]{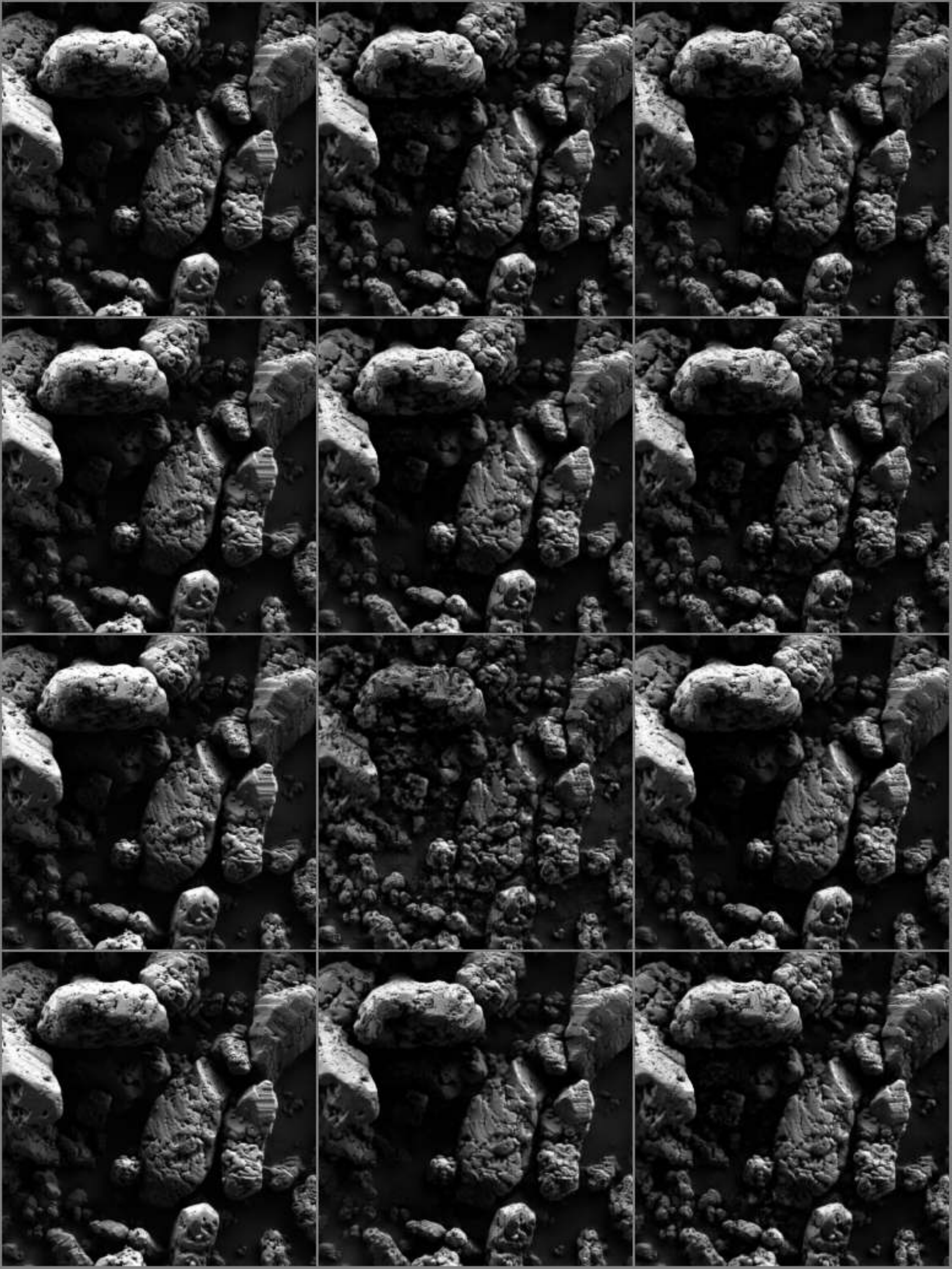}
 \caption{
  Additional examples of GAN generated images. Similar to the corresponding figure in the main text, the left column is the original SEM image. The middle column images are synthesized to increase the corresponding attributes, whereas the right column is synthesized to decrease the corresponding attribute. The four rows correspond to the four attributes, namely, \emph{porosity}, \emph{polydispersity}, \emph{size}, and \emph{facetness}}
\label{fig:additionalResults_6}
\end{figure*}

\begin{figure*}[!htbp]
\centering
  \includegraphics[width=1.0\linewidth]{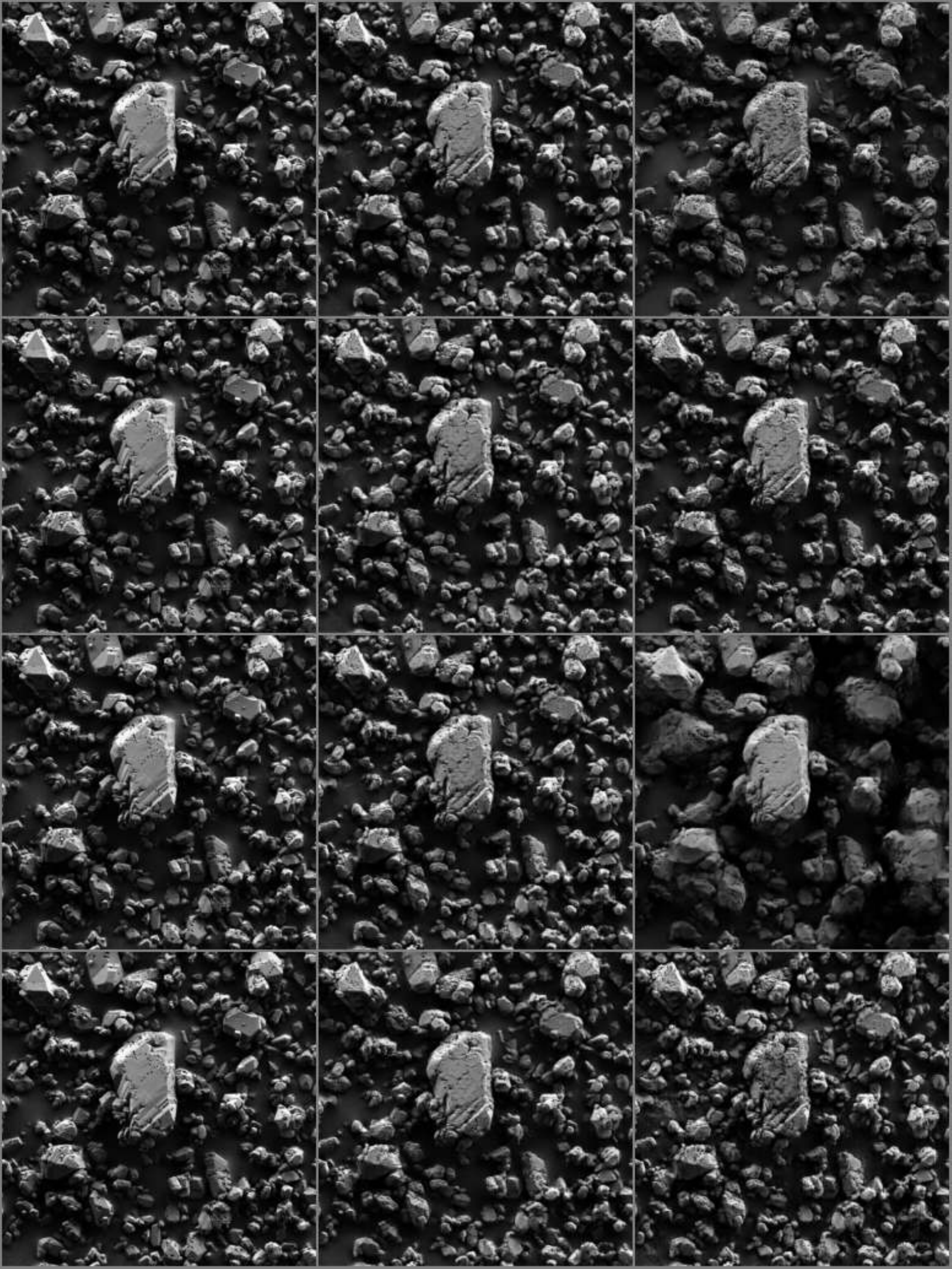}
 \caption{
  Additional examples of GAN generated images. Similar to the corresponding figure in the main text, the left column is the original SEM image. The middle column images are synthesized to increase the corresponding attributes, whereas the right column is synthesized to decrease the corresponding attribute. The four rows correspond to the four attributes, namely, \emph{porosity}, \emph{polydispersity}, \emph{size}, and \emph{facetness}
}
\label{fig:additionalResults_7}
\end{figure*}

To better illustrate the performance our model, we provide additional examples of the GAN-based modifications, as shown in Figure~\ref{fig:additionalResults_1} to Figure~\ref{fig:additionalResults_7}. One thing to note is that many of the modification is reflected in detailed and localized features (e.g., porosity), which may not be obvious at first glance or for people without related background. We advise the reader to zoom in the image to see the details.
For porosity, we can observe a rougher texture with small dips on the crystal in case of increased porosity (rightmost image). For polydispersity, the things to look for is whether smaller crystals are disappearing/appearing between the gaps between larger crystal dependence on decreasing/increasing peak stress prediction.

%% file: SEMpaper.bbl
\begin{thebibliography}{10}

\bibitem{selvaraju2017grad}
Ramprasaath~R Selvaraju, Michael Cogswell, Abhishek Das, Ramakrishna Vedantam,
  Devi Parikh, and Dhruv Batra.
\newblock Grad-cam: Visual explanations from deep networks via gradient-based
  localization.
\newblock In {\em Proceedings of the IEEE International Conference on Computer
  Vision}, pages 618--626, 2017.

\bibitem{gallagher2020predicting}
Brian Gallagher, Matthew Rever, Donald Loveland, T~Nathan Mundhenk, Brock
  Beauchamp, Emily Robertson, Golam~G Jaman, Anna~M Hiszpanski, and T~Yong-Jin
  Han.
\newblock Predicting compressive strength of consolidated molecular solids
  using computer vision and deep learning.
\newblock {\em Materials \& Design}, page 108541, 2020.

\bibitem{ZeilerFergus2014}
Matthew~D Zeiler and Rob Fergus.
\newblock Visualizing and understanding convolutional networks.
\newblock In {\em European conference on computer vision}, pages 818--833.
  Springer, 2014.

\bibitem{bach2015pixel}
Sebastian Bach, Alexander Binder, Gr{\'e}goire Montavon, Frederick Klauschen,
  Klaus-Robert M{\"u}ller, and Wojciech Samek.
\newblock On pixel-wise explanations for non-linear classifier decisions by
  layer-wise relevance propagation.
\newblock {\em PloS one}, 10(7):e0130140, 2015.

\bibitem{goodfellow2014generative}
Ian Goodfellow, Jean Pouget-Abadie, Mehdi Mirza, Bing Xu, David Warde-Farley,
  Sherjil Ozair, Aaron Courville, and Yoshua Bengio.
\newblock Generative adversarial nets.
\newblock In {\em Advances in neural information processing systems}, pages
  2672--2680, 2014.

\bibitem{peterson2017zonal}
JL~Peterson, KD~Humbird, JE~Field, ST~Brandon, SH~Langer, RC~Nora, BK~Spears,
  and PT~Springer.
\newblock Zonal flow generation in inertial confinement fusion implosions.
\newblock {\em Physics of Plasmas}, 24(3):032702, 2017.

\bibitem{anirudh2019improved}
Rushil Anirudh, Jayaraman~J Thiagarajan, Peer-Timo Bremer, and Brian~K Spears.
\newblock Improved surrogates in inertial confinement fusion with manifold and
  cycle consistencies.
\newblock {\em arXiv preprint arXiv:1912.08113}, 2019.

\bibitem{webb2018deep}
Sarah Webb.
\newblock Deep learning for biology.
\newblock {\em Nature}, 554(7693), 2018.

\bibitem{butler2018machine}
Keith~T Butler, Daniel~W Davies, Hugh Cartwright, Olexandr Isayev, and Aron
  Walsh.
\newblock Machine learning for molecular and materials science.
\newblock {\em Nature}, 559(7715):547, 2018.

\bibitem{reichstein2019deep}
Markus Reichstein, Gustau Camps-Valls, Bjorn Stevens, Martin Jung, Joachim
  Denzler, Nuno Carvalhais, et~al.
\newblock Deep learning and process understanding for data-driven earth system
  science.
\newblock {\em Nature}, 566(7743):195--204, 2019.

\bibitem{kurth2018exascale}
Thorsten Kurth, Sean Treichler, Joshua Romero, Mayur Mudigonda, Nathan Luehr,
  Everett Phillips, Ankur Mahesh, Michael Matheson, Jack Deslippe, Massimiliano
  Fatica, et~al.
\newblock Exascale deep learning for climate analytics.
\newblock In {\em SC18: International Conference for High Performance
  Computing, Networking, Storage and Analysis}, pages 649--660. IEEE, 2018.

\bibitem{baldi2001bioinformatics}
Pierre Baldi, S{\o}ren Brunak, and Francis Bach.
\newblock {\em Bioinformatics: the machine learning approach}.
\newblock MIT press, 2001.

\bibitem{SimonyanVedaldiZisserman2013}
Karen Simonyan, Andrea Vedaldi, and Andrew Zisserman.
\newblock Deep inside convolutional networks: Visualising image classification
  models and saliency maps.
\newblock {\em arXiv preprint arXiv:1312.6034}, 2013.

\bibitem{YosinskiCluneNguyen2015}
Jason Yosinski, Jeff Clune, Anh Nguyen, Thomas Fuchs, and Hod Lipson.
\newblock Understanding neural networks through deep visualization.
\newblock {\em arXiv preprint arXiv:1506.06579}, 2015.

\bibitem{lapuschkin2019unmasking}
Sebastian Lapuschkin, Stephan W{\"a}ldchen, Alexander Binder, Gr{\'e}goire
  Montavon, Wojciech Samek, and Klaus-Robert M{\"u}ller.
\newblock Unmasking clever hans predictors and assessing what machines really
  learn.
\newblock {\em Nature communications}, 10(1):1096, 2019.

\bibitem{kailkhura2019reliable}
Bhavya Kailkhura, Brian Gallagher, Sookyung Kim, Anna Hiszpanski, and
  T~Yong-Jin Han.
\newblock Reliable and explainable machine-learning methods for accelerated
  material discovery.
\newblock {\em npj Computational Materials}, 5(1):1--9, 2019.

\bibitem{xie2020explainable}
Ning Xie, Gabrielle Ras, Marcel van Gerven, and Derek Doran.
\newblock Explainable deep learning: A field guide for the uninitiated.
\newblock {\em arXiv preprint arXiv:2004.14545}, 2020.

\bibitem{RibeiroSinghGuestrin2016}
Marco~Tulio Ribeiro, Sameer Singh, and Carlos Guestrin.
\newblock Why should i trust you?: Explaining the predictions of any
  classifier.
\newblock In {\em Proceedings of the 22nd ACM SIGKDD International Conference
  on Knowledge Discovery and Data Mining}, pages 1135--1144. ACM, 2016.

\bibitem{KrausePererNg2016}
Josua Krause, Adam Perer, and Kenney Ng.
\newblock Interacting with predictions: Visual inspection of black-box machine
  learning models.
\newblock In {\em Proceedings of the 2016 CHI Conference on Human Factors in
  Computing Systems}, pages 5686--5697. ACM, 2016.

\bibitem{LundbergLee2017}
Scott~M Lundberg and Su-In Lee.
\newblock A unified approach to interpreting model predictions.
\newblock In {\em Advances in Neural Information Processing Systems}, pages
  4768--4777, 2017.

\bibitem{kusner2017counterfactual}
Matt~J Kusner, Joshua Loftus, Chris Russell, and Ricardo Silva.
\newblock Counterfactual fairness.
\newblock In {\em Advances in Neural Information Processing Systems}, pages
  4066--4076, 2017.

\bibitem{narendra2018explaining}
Tanmayee Narendra, Anush Sankaran, Deepak Vijaykeerthy, and Senthil Mani.
\newblock Explaining deep learning models using causal inference.
\newblock {\em arXiv preprint arXiv:1811.04376}, 2018.

\bibitem{Goyal2019}
Yash Goyal, Ziyan Wu, Jan Ernst, Dhruv Batra, Devi Parikh, and Stefan Lee.
\newblock Counterfactual visual explanations.
\newblock {\em arXiv preprint arXiv:1904.07451}, 2019.

\bibitem{anne2018grounding}
Lisa Anne~Hendricks, Ronghang Hu, Trevor Darrell, and Zeynep Akata.
\newblock Grounding visual explanations.
\newblock In {\em Proceedings of the European Conference on Computer Vision
  (ECCV)}, pages 264--279, 2018.

\bibitem{liu2019generative}
Shusen Liu, Bhavya Kailkhura, Donald Loveland, and Yong Han.
\newblock Generative counterfactual introspection for explainable deep
  learning.
\newblock {\em arXiv preprint arXiv:1907.03077}, 2019.

\bibitem{decost2017characterizing}
Brian~L DeCost and Elizabeth~A Holm.
\newblock Characterizing powder materials using keypoint-based computer vision
  methods.
\newblock {\em Computational Materials Science}, 126:438--445, 2017.

\bibitem{webel2018new}
Johannes Webel, Jessica Gola, Dominik Britz, and Frank M{\"u}cklich.
\newblock A new analysis approach based on haralick texture features for the
  characterization of microstructure on the example of low-alloy steels.
\newblock {\em Materials Characterization}, 144:584--596, 2018.

\bibitem{zagoruyko2016wide}
Sergey Zagoruyko and Nikos Komodakis.
\newblock Wide residual networks.
\newblock {\em arXiv preprint arXiv:1605.07146}, 2016.

\bibitem{brock2018large}
Andrew Brock, Jeff Donahue, and Karen Simonyan.
\newblock Large scale gan training for high fidelity natural image synthesis.
\newblock {\em arXiv preprint arXiv:1809.11096}, 2018.

\bibitem{karras2019style}
Tero Karras, Samuli Laine, and Timo Aila.
\newblock A style-based generator architecture for generative adversarial
  networks.
\newblock In {\em Proceedings of the IEEE Conference on Computer Vision and
  Pattern Recognition}, pages 4401--4410, 2019.

\bibitem{mirza2014conditional}
Mehdi Mirza and Simon Osindero.
\newblock Conditional generative adversarial nets.
\newblock {\em arXiv preprint arXiv:1411.1784}, 2014.

\bibitem{attGAN}
Zhenliang He, Wangmeng Zuo, Meina Kan, Shiguang Shan, and Xilin Chen.
\newblock Attgan: Facial attribute editing by only changing what you want.
\newblock {\em IEEE Transactions on Image Processing}, 2019.

\bibitem{potter2006methods}
Kristin Potter, Hans Hagen, Andreas Kerren, and Peter Dannenmann.
\newblock Methods for presenting statistical information: The box plot.
\newblock {\em Visualization of large and unstructured data sets}, 4:97--106,
  2006.

\bibitem{sohn1968effect}
Hong~Yong Sohn and C~Moreland.
\newblock The effect of particle size distribution on packing density.
\newblock {\em The Canadian Journal of Chemical Engineering}, 46(3):162--167,
  1968.

\bibitem{mallick2019deep}
Ankur Mallick, Chaitanya Dwivedi, Bhavya Kailkhura, Gauri Joshi, and T~Han.
\newblock Deep probabilistic kernels for sample-efficient learning.
\newblock {\em arXiv preprint arXiv:1910.05858}, 2019.

\bibitem{kingma2014adam}
Diederik~P Kingma and Jimmy Ba.
\newblock Adam: A method for stochastic optimization.
\newblock {\em arXiv preprint arXiv:1412.6980}, 2014.

\end{thebibliography}
